\documentclass[lettersize,journal]{IEEEtran}
\usepackage{amsmath,amsfonts,amssymb,amsthm}
\usepackage{algorithmic}
\usepackage{algorithm}
\usepackage{multicol}
\usepackage{array}
\usepackage[caption=false,font=normalsize,labelfont=sf,textfont=sf]{subfig}
\usepackage{textcomp}
\usepackage{stfloats}
\usepackage{url}
\usepackage{verbatim}
\usepackage{graphicx}
\usepackage{cite}
\usepackage{mathrsfs}
\usepackage{longtable}
\usepackage{booktabs}
\usepackage{float}
\usepackage{makecell}
\usepackage{bigstrut}
\usepackage{enumitem}
\usepackage{bm} 
\usepackage{threeparttable}
\usepackage{multirow}
\usepackage{booktabs}
\usepackage{pifont} 
\usepackage{listings}
\newcommand{\cmark}{\ding{51}} 
\newcommand{\xmark}{\ding{55}} 

\usepackage{caption}
\usepackage[colorlinks, linkcolor=blue, citecolor=blue, urlcolor=blue]{hyperref}
\usepackage{booktabs,siunitx,tabularx,caption}
\newcolumntype{L}[1]{>{\raggedright\arraybackslash}p{#1}}
\newcolumntype{C}[1]{>{\centering\arraybackslash}p{#1}}
\newcolumntype{R}[1]{>{\raggedleft\arraybackslash}p{#1}}

\begin{document}

\title{Decision-Focused Continual Learning for Seaport Power–Logistics Scheduling: Generalization across Varying Tasks}

\author{
  Chuanqing Pu,~\IEEEmembership{Student Member,~IEEE,}
  Feilong Fan,~\IEEEmembership{Member,~IEEE,}~
  Nengling Tai,~\IEEEmembership{Senior Member,~IEEE,}~\\
  Yan Xu,~\IEEEmembership{Senior Member,~IEEE,}
  Wentao Huang,~\IEEEmembership{Senior Member,~IEEE,}~and~
  Honglin Wen,~\IEEEmembership{Member,~IEEE}~

  \thanks{Chuanqing Pu, Feilong Fan and Nengling Tai are with the College of Smart Energy, Shanghai Jiao Tong University, Shanghai 201100, China (email: sashabanks@sjtu.edu.cn; feilongfan@sjtu.edu.cn; nltai@sjtu.edu.cn). Yan Xu is with the School of Electrical and Electronic Engineering, Nanyang Technological University, Singapore (email: xuyan@ntu.edu.sg). Wentao Huang and Honglin Wen are with the School of Electrical Engineering, Shanghai Jiao Tong University, Shanghai 200240, China (email: hwt8989@sjtu.edu.cn). Honglin Wen is also with the Dyson School of Design Engineering, Imperial College London, United Kingdom (email: linlin00@sjtu.edu.cn).}
  \thanks{This work was supported by National Natural Science Foundation of China under Grant 52407125, and National Key R\&D Project (No. 2024YFB4206500). \textit{(Corresponding author: Feilong Fan.)}}%
  \vspace{-2em}
}

\maketitle


\begin{abstract}
Power–logistics scheduling in modern seaports typically follows a predict-then-optimize pipeline. To enhance the decision quality of predictions, decision-focused learning has been proposed, which aligns the training of forecasting models with downstream decision outcomes. However, this end-to-end design inherently restricts the value of forecasting models to a specific task structure and therefore generalizes poorly to evolving tasks induced by varying vessel arrivals. We address this gap with a decision-focused continual learning framework that adapts online to a stream of scheduling tasks. Specifically, we introduce Fisher-information-based regularization to enhance cross-task generalization by preserving parameters critical to prior tasks. A differentiable convex surrogate is also developed to stabilize gradient backpropagation. The proposed approach enables learning a decision-aligned forecasting model across a varying task stream with sustainable long-term computational and memory requirements. Experiments calibrated to Jurong Port show improved decision performance and cross-task generalization over existing methods, together with reduced computational cost and a bounded memory footprint.
\end{abstract}

\vspace{-1em}

\section{Introduction}
\subsection{Background and Motivation}
\IEEEPARstart{W}ith the accelerated electrification of logistics operations, modern seaports are progressively transforming into power–logistics coupled systems (PLS) characterized by high renewable penetration and flexible logistics loads \cite{PLS_1}. In this context, extensive research has been devoted to enhancing the operational benefits of PLS through advanced forecasting and scheduling techniques \cite{PLS_sh_1,PLS_sh_2,PLS_sh_3}. Specifically, forecasting models capture the stochastic dynamics of net load demand and electricity prices, while PLS schedulers leverage these forecasts to co-optimize energy procurement and logistics operations. This sequential workflow can be formulated as a \emph{predict–then–optimize} pipeline \cite{SPO} that mirrors real-world industrial practice in seaport PLS operations.

\begin{figure*}[!ht]
    \centering
    \includegraphics[width=1\textwidth]{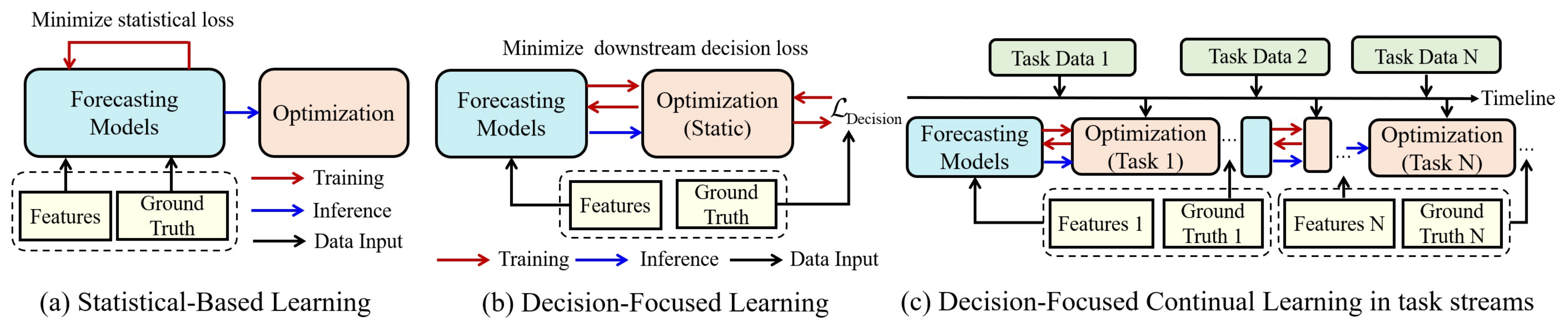}
    \caption{Illustration of different learning paradigms for predict-then-optimize pipelines.}
    \label{fig:intro}
\end{figure*}

In prevailing predict–then–optimize pipelines, forecasting models are typically trained to minimize statistical losses (e.g. mean squared error, MSE), as shown in Fig.~\ref{fig:intro}(a). However, the decision quality achieved through downstream optimization depends not only on the statistically pointwise accuracy but also on the direction of bias and the temporal structure of the forecast errors \cite{DFL-RL}. For instance, when a PLS operator forecasts the day-ahead net-load profile, overestimation may lead to excessive energy procurement and trigger downward regulation costs, whereas underestimation causes real-time shortages that trigger upward regulation \cite{PLS_sh_4}. The regulation cost is inherently asymmetric under different imbalance settlement mechanisms and also coupled with the day-ahead price profile. These asymmetric and nonlinear effects of forecast errors on scheduling are poorly captured by symmetric, linear statistical losses such as MSE or mean absolute error (MAE). 

To better align forecasting with downstream operational value, decision-focused learning (DFL) evaluates forecast quality through its impact on the underlying optimization task rather than through pointwise prediction errors alone \cite{DFL-old}. As depicted in Fig.~\ref{fig:intro}(b), DFL integrates forecasting and optimization in an end-to-end pipeline and trains the forecasting model directly with a decision-related loss.

While this end-to-end paradigm can improve decision performance for a given task, it also inherently confines the forecasting model's decision quality to a fixed and predetermined optimization task. In contrast, PLS scheduling tasks evolve with vessel arrivals, since logistics decisions must be made conditional on arrival-dependent exogenous parameters \cite{PLS_sh_4}. Each new arrival configuration introduces a distinct task instance characterized by different numbers of vessels, cargo profiles, and time-coupled constraints. Although these tasks share similar high-level formulations, they differ in the dimensions of decision variables and the geometry of feasible regions. As a result, the ground-truth objectives and loss landscapes of DFL vary across configurations. This is the key distinction between PLS operation and conventional DFL settings in power systems. Motivated by this characteristic, this paper investigates the following question: \emph{How can a forecasting model be trained to remain decision-effective across an evolving stream of tasks?}

\subsection{Related Work and Research Gap}
\textbf{Decision-focused learning for energy applications.}
Recent studies have substantially advanced DFL methodologies and applications in energy systems. Representative examples include: (i) value-/cost-oriented forecasting and trading, which explicitly bridges forecasting objectives with market settlement and value metrics \cite{li2016toward,georges2020integrated,akylas2022trees,DFL-Inertia}; (ii) end-to-end DFL pipelines enabled by implicit differentiation through convex optimization layers, which tightly integrate forecasting with economic dispatch or storage arbitrage decisions \cite{pu2025hybrid,han2021task-based,wahdany2023more,sang2022safety,ruben2025vof,xubolun2025} and (iii)  DFL for tasks involving discrete decisions and non-convex objectives (e.g. unit commitment or multi-energy dispatch), where surrogate models, subgradient constructions, or perturbation-based estimators are adopted to enable gradient-based training \cite{munoz2022bilevel,sang2022electricity,chen2021feature,xie2025predict,yangze2024load,wancan2025weight}. As summarized in Table~\ref{tab:litreview}, these efforts span diverse problem types and gradient mechanisms, yet they are predominantly task-specific and assume a known task structure. This premise becomes restrictive in the settings with evolving tasks, where the learned representations may not generalize reliably to new optimization task configurations. 

\textbf{Decision-focused fine-tuning.} Decision-focused fine-tuning has been explored as a pragmatic alternative, which adapts a pre-trained predictor to a new decision environment using limited data when full end-to-end retraining is costly \cite{yang1, yang2}. Its primary emphasis is to adapt the model to the currently observed task distribution, rather than to maintain decision effectiveness over a stream of heterogeneous tasks. As a result, when task configurations evolve sequentially, fine-tuning provides no explicit mechanism to preserve decision-relevant knowledge acquired from earlier tasks or to consolidate shared structure across tasks. However, task variations driven by seaport vessel arrivals call for a learning paradigm that can adapt to newly realized distributions while preserving decision quality on previously encountered tasks.

\textbf{Multi-task and multi-objective DFL.}
Recent work also explores multi-task and multi-objective DFL variants, in which forecasting models are trained to minimize the average loss across a predefined set of tasks or objectives \cite{DFMTL,DFMOL}. Another line of work pretrains a surrogate network to capture the joint effect of forecasts and exogenous task parameters on operational cost \cite{DFL-surrogate}. Robust task-aware end-to-end learning has also been studied to improve robustness against bounded uncertainty in input features and unpredictable optimization parameters \cite{xu2024e2eat}. Although these extensions improve cross-task generalization or robustness, they still presuppose prior knowledge of possible task structures and increase the demand for large and diverse offline datasets. In PLS operation, future vessel-arrival plans are uncertain, therefore the task configurations that will be encountered cannot be fully specified in advance \cite{PLS_1}. Evolving seaport vessel-arrival conditions may induce unseen combinations of fleet size, cargo profiles, and time-coupled constraints. Such variations alter not only input parameters, but also the dimension of decision variables and the geometry of the feasible region. Consequently, these structural shifts are difficult to characterize using compact uncertainty sets or to cover exhaustively through offline training. By contrast, the proposed online adaptation mechanism is intended for the open-ended setting where new task structures are revealed sequentially rather than known a priori.

\textbf{Continual learning.} Online continual learning (CL) offers a path for long-term adaptation with bounded computational and memory resources, yet directly applying generic CL paradigms to DFL faces two key obstacles. First, replay-based CL can be computationally prohibitive in decision-focused settings because each replay step may require repeated evaluations and differentiation of the embedded optimization layer \cite{CL_vs_MTL,CL_replay}. Second, parameter-isolation \cite{rusu2016progressive} or dynamically expanding architectures \cite{mallya2018packnet} may avoid forgetting but typically incur model growth and/or require reliable task identifiers, which conflicts with the open-ended and continuously evolving task landscape in PLS.

\begin{table*}[!ht]
\centering
\caption{Representative studies on Decision-Focused Learning in energy and transportation systems}
\label{tab:litreview}
\renewcommand{\arraystretch}{1.18}
\setlength{\tabcolsep}{5.8pt}
\begin{threeparttable}
\begin{tabularx}{\textwidth}{p{2.5cm} p{3.8cm} p{1.8cm} c c c c c}
\hline
\hline
\textbf{Related Work} & \textbf{Task Domain} & \textbf{Problem Type}& \textbf{Gradient-based} & \textbf{Fine-tuning} & \textbf{Multi-task} & \textbf{Task Adaptive} \\
\midrule

\makecell[l]{
\cite{li2016toward};
\cite{georges2020integrated};
\cite{akylas2022trees};
\cite{DFL-Inertia}
}
& Energy trading & LP/MIP & \xmark & \xmark & \xmark & \xmark \\

\midrule
\makecell[l]{
\cite{pu2025hybrid};
\cite{han2021task-based};
\cite{wahdany2023more};
\cite{sang2022safety}};
\cite{ruben2025vof};
\cite{xubolun2025}
& \makecell[l]{Economic dispatch;\\ Energy trading} & LP/QP/MIP 
& \makecell{Analytical} & \xmark & \xmark & \xmark\\

\midrule
\makecell[l]{\cite{munoz2022bilevel},\cite{sang2022electricity}; \cite{chen2021feature};\cite{xie2025predict};\\\cite{yangze2024load};\cite{wancan2025weight}}
& \makecell[l]{Economic dispatch;\\ Unit commitment}
& SOCP/MIP
& \makecell{Empirical (SPO+/surrogate)} & \xmark & \xmark & \xmark \\

\midrule
\makecell[l]{\cite{yang1};\cite{yang2}}
& \makecell[l]{Ride-hailing subsidy allocation; \\ Transportation resource scheduling}
& LP/MIP
& \makecell{Empirical (SPO+)} & \cmark & \xmark & \xmark \\

\midrule
\makecell[l]{\cite{DFMTL};\cite{DFMOL};\cite{DFL-surrogate};\cite{xu2024e2eat}}
& \makecell[l]{Travelling Salesman;\\ Robust Portfolio; \\  Economic Dispatch}
& MIP
& \makecell{Empirical} & \xmark & \cmark & \xmark \\

\midrule
\makecell[l]{\textbf{Proposed}}
& \makecell[l]{Power–Logistics Scheduling}
& MIP
& \makecell{Empirical (Surrogate)} & \cmark & \cmark & \cmark\\

\hline
\hline
\end{tabularx}

\vspace{3pt}
\begin{tablenotes}[flushleft]
\footnotesize
\item
\textbf{Problem Type}: LP, QP, SOCP and MIP denote linear, quadratic, second-order cone, and mixed-integer programming, respectively.
\end{tablenotes}
\end{threeparttable}
\end{table*}

\subsection{Our Contributions}
To address these challenges, we reformulate conventional static DFL into a decision-focused continual learning (DFCL) paradigm, as illustrated in Fig.~\ref{fig:intro}(c). DFCL sequentially learns over a stream of PLS tasks in an online manner. Inspired by Elastic Weight Consolidation (EWC) \cite{CL_regularization}, we derive a regularization method that constrains updates to parameters critical for past decision mappings. This design allows the forecasting model to preserve decision-effective information accumulated from previous tasks while adapting to new task conditions. Compared to replay-based continual learning, which requires frequent re-optimization over historical data \cite{CL_replay}, the proposed approach preserves cross-task generalization without repeated evaluations of the embedded optimization layer. Moreover, the embedded non-differentiability of the PLS scheduling problem is addressed through a differentiable surrogate that locally convexifies the problem during back-propagation. The remaining convex structure can then be efficiently differentiated through implicit differentiation \cite{OptNet}. In a representative benchmark inspired by Jurong Port, conventional task-specific DFL settings indeed fail to sustain decision performance when sequentially adapted to task streams. The proposed DFCL consistently outperforms statistic-based learning (SBL) and existing DFL baselines. The main contributions of this work are summarized as follows:

\begin{itemize}
  \item A novel task-adaptive predict-then-optimize framework is proposed for PLS scheduling, which enables continually learning a decision-aligned forecasting model while preserving generalization over an unbounded task stream.
  
  \item A Fisher-information-based regularization method is proposed to enhance cross-task generalization over the task stream. This avoids prohibitively high computational overhead from frequent retraining on past tasks.
  
  \item An end-to-end learning pipeline is developed, which employs a differentiable surrogate to convexify the embedded mixed-integer PLS scheduling problem and stabilize gradient back-propagation by implicit differentiation.
  
\end{itemize}

The rest of this paper is organized as follows. Section \ref{sec:2} presents the predict-then-optimize pipeline of PLS. Section \ref{sec:3} formulates the DFCL problem for the PLS task stream. Section \ref{sec:4} introduces the differentiable end-to-end learning pipeline to implement DFCL. Section \ref{sec:5} reports the experimental setup and results. Section \ref{sec:6} concludes the paper.

\section{Predict-then-Optimize Pipeline in PLS}
\label{sec:2}
This study considers a seaport PLS operating as a price-taking industrial consumer in a retail electricity market. The energy charge is assumed to be indexed to the wholesale day-ahead price signal, treated as exogenous \cite{EurelectricHedging2024}, and deviations from the day-ahead schedule are settled via a single-price imbalance mechanism \cite{ACERISH2020}. Accordingly, we abstract away from strategic price bidding and market-clearing modeling, and instead focus on how forecast errors translate into downstream co-scheduling outcomes and imbalance costs under exogenous price signals.

The PLS comprises photovoltaics (PV) generation, an energy storage system (ESS), uncontrollable residential loads and a logistics subsystem including vessel berthing, quay-crane operations, and shore power supply. The detailed structural illustration can be found in Appendix~\ref{sec:appendixA}. Below we mainly describe the operational workflow and problem formulation.

\begin{figure}[!ht]
    \centering
    \includegraphics[width=0.48\textwidth]{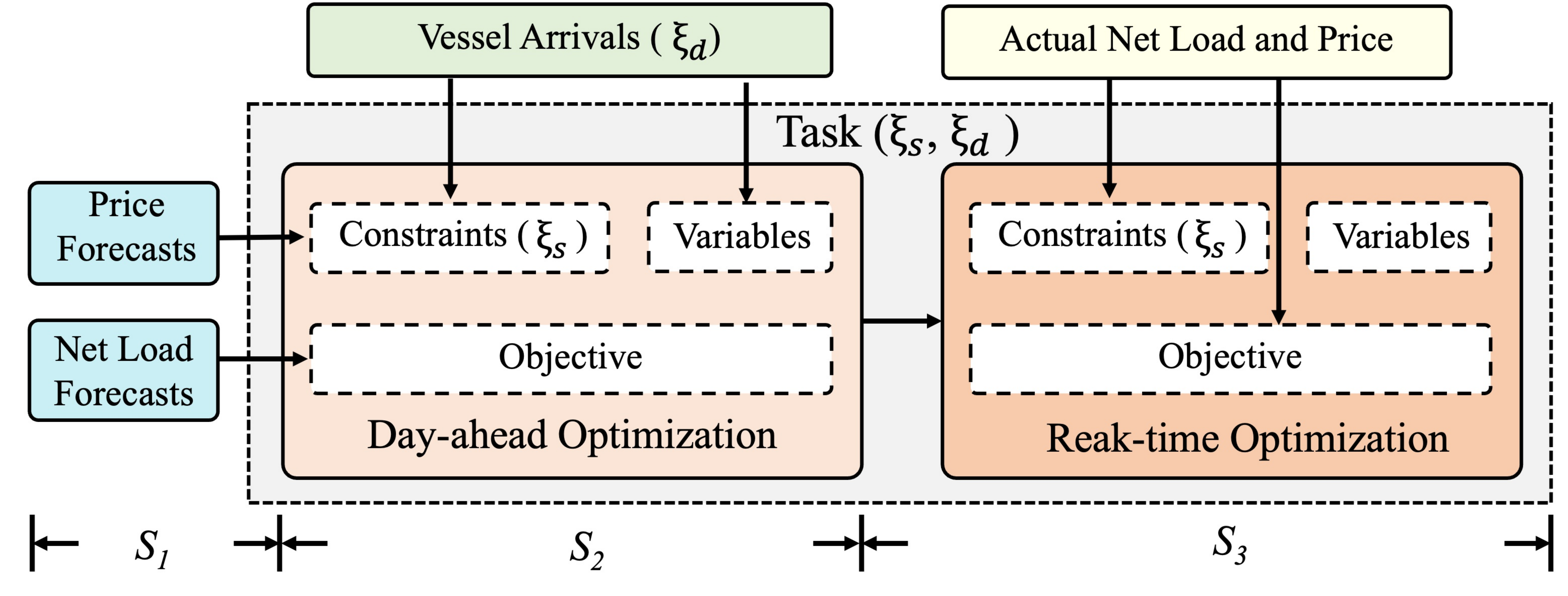}
    \caption{Illustration of the predict-then-optimize pipeline in seaport PLS scheduling.}
    \label{fig:task}
\end{figure}

The operation of the seaport PLS is modeled as a sequential \textit{predict–then–optimize} pipeline that mirrors real industrial practice. It consists of three consecutive stages: 
\textit{(i) Day-ahead forecasting ($\mathcal{S}_1$)}, 
\textit{(ii) Day-ahead scheduling and bidding ($\mathcal{S}_2$)}, and \textit{(iii) Real-time corrective scheduling with limited recourse ($\mathcal{S}_3$)}, as illustrated in Fig.~\ref{fig:task}. Each stage is parameterized by the outputs of the preceding stages. The final performance of the entire pipeline is evaluated by the actual operating cost incurred in real-time scheduling. The following subsections will introduce the formulation of each stage in detail.

\subsection{Day-ahead Forecasting}
The stage $\mathcal{S}_1$ predicts two types of exogenous input for the following stages over a planning horizon of $T$ time intervals: the \textit{market electricity price} $\hat{\boldsymbol{\pi}}\!\in\!\mathbb{R}^{T},$ and the \textit{net load} $\hat{\mathbf{p}}\!\in\!\mathbb{R}^{T}$, defined as the uncontrollable demand minus PV generation within the port area. The predicted $\hat{\boldsymbol{\pi}}$ and $\hat{\mathbf{p}}$ serve as the basis for $\mathcal{S}_2$. 
Let $\boldsymbol{\phi}_{\mathrm{p}}$ and $\boldsymbol{\phi}_{\mathrm{l}}$ denote the contextual feature vectors for price and load forecasting, respectively, and $\boldsymbol{\theta}_{\mathrm{p}}$, $\boldsymbol{\theta}_{\mathrm{l}}$ the corresponding model parameters. The forecasting models are formulated as
\begin{equation}
\label{eq:forecasting}
\hat{\boldsymbol{\pi}} = g_{\pi}(\boldsymbol{\phi}_{\text{p}};\boldsymbol{\theta}_\text{p}) ,\quad 
\hat{\mathbf{p}} = g_{\text{l}}(\boldsymbol{\phi}_{\text{l}};\boldsymbol{\theta}_\text{l}),
\end{equation}
where $g_{\pi}(\cdot)$ and $g_{\text{l}}(\cdot)$ denote the forecasting functions for price and load, respectively.

\subsection{Day-ahead Scheduling}

The stage $\mathcal{S}_2$ parameterizes an optimization model not only based on the forecasts from $\mathcal{S}_1$ and static endogenous parameters $\boldsymbol{\xi}_s$, but also requires dynamic exogenous parameters $\boldsymbol{\xi}_d$ to describe the task instance. The static parameters $\boldsymbol{\xi}_s$ refer to the fixed attributes within the PLS, such as ESS capacity, logistics infrastructure, and operational limits, which remain unchanged over long time scales and are shared across different task instances. The dynamic parameters $\boldsymbol{\xi}_d$ refer to the variable vessel arrival schedule that is only revealed at the day-ahead scheduling stage. The dynamic parameter vector $\boldsymbol{\xi}_d$ includes information such as the number of arriving vessels, cargo volumes, and schedules. This information jointly determines the structure and dimension of the decision variables and constraints in the day-ahead scheduling model.

The day-ahead schedule specifies the charging/discharging power of the ESS ($\mathbf{P}_{\mathrm{ES.ch}},\mathbf{P}_{\mathrm{ES.dch}}\!\in\!\mathbb{R}^{T}$), the shore-power supply $\mathbf{P}_{\mathrm{s}}\!\in\!\mathbb{R}^{T}$ to berthed vessels and the quay-crane power $\mathbf{P}_{\mathrm{QC}}\!\in\!\mathbb{R}^{T}$, together with binary logistics indicators $\mathbf{V}\! \in \!\mathbb{R}^{T \times J}$ that represent vessel-berth assignment decisions, where $J$ is the number of total vessel arrivals. As a price-taker in the retail market, the PLS needs to submit a planned power schedule $\mathbf{P}_{\mathrm{b}}$ to the market as a day-ahead bid. The schedule $\mathbf{P}_{\mathrm{b}}$ is determined by the co-optimization of energy and logistics operations, which is parameterized by the forecasts $(\hat{\boldsymbol{\pi}},\hat{\mathbf{p}})$ and the scheduling configuration. The actual electricity price $\boldsymbol{\pi}$ and realized net load $\mathbf{p}$ are revealed later in $\mathcal{S}_3$, where deviations between the day-ahead bid $\mathbf{P}_{\mathrm{b}}$ and realized consumption $\mathbf{P}_{\mathrm{b}}'$ are settled according to imbalance coefficients $\rho^{+}$ and $\rho^{-}$ for up- and down-regulation, respectively. The rule for imbalance settlement is determined by the retail market and is known to the PLS. 

Given these parameters, let $\mathcal{X}_{\mathrm{ES}}(\boldsymbol{\xi}_s)$ and $\mathcal{X}_{\mathrm{LG}}(\boldsymbol{\xi}_s,\boldsymbol{\xi}_d)$ denote the feasible sets of the ESS and logistics subsystems, respectively. Define the decision vector as
$\mathbf{x}:=[\mathbf{P}_{\mathrm{b}}^\top,\mathbf{\hat{P}}_{\mathrm{b}}^{'\top},\mathbf{P}_{\mathrm{ES.ch}}^\top,\mathbf{P}_{\mathrm{ES.dch}}^\top,\mathbf{P}_{\mathrm{QC}}^\top,\mathbf{P}_{\mathrm{s}}^\top,\mathbf{V}^\top]^{\top}$. The day-ahead scheduling problem can then be formulated as
\begin{subequations}
\label{eq:bidding}
\begin{align}
\min_{\mathbf{x}}\quad 
  & f(\mathbf{P}_{\text{b}},\mathbf{\hat{P}}_{\text{b}}';\hat{\boldsymbol{\pi}})
      = \boldsymbol{\hat{\pi}}^{\top}\mathbf{P}_{\text{b}}
      + \rho^{+}\boldsymbol{\hat{\pi}}^{\top}\Delta\mathbf{P}^{+}
      - \rho^{-}\boldsymbol{\hat{\pi}}^{\top}\Delta\mathbf{P}^{-} \label{eq:bidding1} \\
\text{s.t.}\quad 
    & \bigl[\mathbf{P}_{\text{ES.ch}}^{\top},\mathbf{P}_{\text{ES.dch}}^{\top}\bigr]^{\top} 
      \in \mathcal{X}_{\text{ES}}(\boldsymbol{\xi}_s), \label{eq:bidding2} \\
    & \bigl[\mathbf{P}_{\text{s}}^{\top},\mathbf{P}_{\text{QC}}^{\top},\mathbf{V}^{\top},
      \mathbf{z}(\mathbf{x})\bigr]^{\top} 
      \in \mathcal{X}_{\text{LG}}(\boldsymbol{\xi}_s,\boldsymbol{\xi}_d), \label{eq:bidding3} \\
    & \Delta\mathbf{P}^{+},\;\Delta\mathbf{P}^{-} \ge 0, \label{eq:bidding4} \\
    & \Delta\mathbf{P}^{+}-\Delta\mathbf{P}^{-}
      = \mathbf{\hat{P}}_{\text{b}}' - \mathbf{P}_{\text{b}}, \label{eq:bidding5} \\
    & \mathbf{\hat{P}}_{\text{b}}'
      = \mathbf{P}_{\text{s}} + \mathbf{P}_{\text{QC}}
        - \mathbf{P}_{\text{ES.dch}} + \mathbf{P}_{\text{ES.ch}} + \hat{\mathbf{p}},
        \label{eq:bidding6}
\end{align}
\end{subequations}
where the variable $\mathbf{\hat{P}}_{\text{b}}'$ denotes the estimated actual power consumption according to the day-ahead schedule and forecasts, $\Delta\mathbf{P}^{+}$ and $\Delta\mathbf{P}^{-}$ represent the upward and downward deviations from the bid, respectively, and $\mathbf{z}(\mathbf{x})$ represents the remaining intermediate variables determined by $\mathbf{x}$. The first term of \eqref{eq:bidding1} represents the ex-ante cost on the predicted price, and the second and third terms account for penalties associated with deviations from the bid in actual power consumption. \eqref{eq:bidding4} and \eqref{eq:bidding5} are used to linearize the cost function, and \eqref{eq:bidding6} represents the power balance constraint. Detailed formulations of the ESS and logistics feasible sets $\mathcal{X}_{\mathrm{ES}}(\cdot)$ and $\mathcal{X}_{\mathrm{LG}}(\cdot)$ can be found in Appendix~\ref{sec:appendixA}.

\subsection{Real-time Scheduling with Limited Recourse}
Real-time scheduling aims to determine the actual power demand of the PLS through limited corrective actions after the day-ahead plan is fixed. In the present formulation, berth allocation, quay-crane assignment, vessel service sequencing, and ESS charging/discharging are treated as day-ahead commitments, because these decisions are strongly coupled over time and are tied to berth windows, labor coordination, and operational safety requirements. By contrast, the shore power supply and the realized grid withdrawal can still be adjusted in real time based on ultra-short-term updates. This real-time stage is introduced to evaluate how the day-ahead bid performs in operation once the committed logistics and ESS decisions are carried forward.

Given that ultra-short-term updates have a very short lead time relative to actual scheduling actions, we use them as proxies for realized conditions when evaluating the decision value of the day-ahead forecasts \cite{DFL-ED2,yangze2024load}. Accordingly, the real-time scheduling can be formulated as the following equivalent problem
\begin{subequations}
\label{eq:realtime}
\begin{align}
\min_{\mathbf{P}_\text{b}^{'},\,\mathbf{P}_\text{s}} \quad & f(\mathbf{P}_\text{b}^{\ast}(\boldsymbol{\hat{\pi}},\mathbf{\hat{p}}),\mathbf{P}_\text{b}^{'};\boldsymbol{\pi}) \label{eq:realtime1} \\
\text{s.t.}\quad
& [\mathbf{P}_\text{s}^{\top},\mathbf{P}_\text{QC}^{\ast \top},\mathbf{V}^{\ast\top}]^\top \in \mathcal{X}_\text{LG}(\boldsymbol{\xi}_s,\boldsymbol{\xi}_d), \label{eq:realtime2}\\
& \mathbf{P}_b^{'} = \mathbf{P}_\text{s}+\mathbf{P}_\text{QC}^{\ast}-\mathbf{P}_\text{ES.dch}^{\ast}+\mathbf{P}_\text{ES.ch}^{\ast}+\mathbf{p}, \label{eq:realtime3}
\end{align}
\end{subequations}
where $\mathbf{P}_\text{b}^{\ast}(\boldsymbol{\hat{\pi}},\mathbf{\hat{p}})$, $\mathbf{P}_\text{ES.dch}^{\ast}$, $\mathbf{P}_\text{ES.ch}^{\ast}$, $\mathbf{P}_\text{QC}^{\ast}$, and $\mathbf{V}^{\ast}$ are the optimal solutions obtained from \eqref{eq:bidding}. The actual net load $\mathbf{p}$ is used in place of the predicted net load $\mathbf{\hat{p}}$ in \eqref{eq:realtime3}.

\section{Decision-Focused Continual Learning in PLS}
\label{sec:3}

\subsection{Decision-Focused Learning Formulation}
Through the above predict-then-optimize pipeline, the final daily operating cost of PLS can be calculated as $f(\mathbf{P}_\text{b}^{\ast}(\boldsymbol{\hat{\pi}},\mathbf{\hat{p}}),\mathbf{P}_\text{b}^{'\ast}(\boldsymbol{\hat{\pi}},\mathbf{\hat{p}}),\boldsymbol{\pi})$, where $\mathbf{P}_\text{b}^{'\ast}(\boldsymbol{\hat{\pi}},\mathbf{\hat{p}})$ is the optimal solution of \eqref{eq:realtime} given the day-ahead decisions based on the predictions $\boldsymbol{\hat{\pi}}$ and $\mathbf{\hat{p}}$. For readability, the shorthand $\mathbf{P}_\text{b}^{'\ast}(\boldsymbol{\hat{\pi}},\mathbf{\hat{p}})$ suppresses the dependence of the real-time optimizer on the realized quantities $(\boldsymbol{\pi},\mathbf{p})$ and on the fixed day-ahead decisions induced by $(\boldsymbol{\hat{\pi}},\mathbf{\hat{p}})$.

To evaluate the decision-making value of forecasting models, we define the regret function as
\begin{equation}
  \label{eq:regret}
  \begin{aligned}
    \mathcal{L}_\text{reg}(\boldsymbol{\hat{\pi}},\boldsymbol{\pi},\mathbf{\hat{p}},\mathbf{p})= & f(\mathbf{P}_\text{b}^{\ast}(\boldsymbol{\hat{\pi}},\mathbf{\hat{p}}),\mathbf{P}_\text{b}^{'\ast}(\boldsymbol{\hat{\pi}},\mathbf{\hat{p}});\boldsymbol{\pi}) \\
    & - f(\mathbf{P}_\text{b}^{\ast}(\boldsymbol{\pi},\mathbf{p}),\mathbf{P}_\text{b}^{'\ast}(\boldsymbol{\pi},\mathbf{p});\boldsymbol{\pi}).  
  \end{aligned}
\end{equation}
The first term denotes the actual cost incurred when decisions are based on predicted values, and the second term denotes the cost achievable with perfect foresight. The regret therefore measures the additional cost caused by forecast errors.

Let $n\in\{1,\dots,N\}$ index daily samples.
For each sample $n$, we observe feature vectors for price and net-load forecasting, denoted by $\boldsymbol{\phi}_{\mathrm{p}}^{(n)}$ and $\boldsymbol{\phi}_{\mathrm{l}}^{(n)}$, together with their realized labels $\boldsymbol{\pi}^{(n)}$ and $\mathbf{p}^{(n)}$.
A key modeling point is that $\boldsymbol{\xi}_d^{(k)}$ is exogenous and fixed within a task, while $(\boldsymbol{\phi}_{\mathrm{p}}^{(n)},\boldsymbol{\phi}_{\mathrm{l}}^{(n)},\boldsymbol{\pi}^{(n)},\mathbf{p}^{(n)})$ vary across samples $n$ and can be shared across different tasks.
Accordingly, we define the $k$th task dataset as
\begin{equation}
\mathcal{D}_{k}
=\Big\{\big(\boldsymbol{\phi}_{\mathrm{p}}^{(n)},~\boldsymbol{\phi}_{\mathrm{l}}^{(n)},~\boldsymbol{\pi}^{(n)},~\mathbf{p}^{(n)},~\boldsymbol{\xi}_d^{(k)}\big)\Big\}_{n=1}^{N_k}.
\label{eq:task_dataset}
\end{equation}

For a specific task $k$, the objective of DFL is to find the optimal parameters $\boldsymbol{\theta_\text{p}}$ and $\boldsymbol{\theta_\text{l}}$ that minimize the regret, i.e., to identify the parameters that are most beneficial for the entire predict-then-optimize pipeline. This objective can be formulated as a tri-level optimization problem:
\begin{subequations}
\label{eq:DFL}
\begin{align}
\min_{\boldsymbol{\theta_\text{p}},\boldsymbol{\theta_\text{l}}} \quad & \mathbb{E}_{(\boldsymbol{\phi}_\text{p},\boldsymbol{\phi}_\text{l},\boldsymbol{\pi},\mathbf{p})\sim \mathcal{D}_{k}} \bigl[\mathcal{L}_\text{reg}(\boldsymbol{\hat{\pi}},\boldsymbol{\pi},\mathbf{\hat{p}},\mathbf{p})\bigr] \label{eq:DFL1} \\
\text{s.t.}\quad
& \mathbf{P}_\text{b}^{\ast}(\boldsymbol{\hat{\pi}},\mathbf{\hat{p}}),\mathbf{P}_\text{ES.dch}^{\ast},\mathbf{P}_\text{ES.ch}^{\ast},\mathbf{P}_\text{QC}^{\ast},\mathbf{V}^{\ast} \\
& = \arg \min_{\mathbf{x}} f(\mathbf{P}_\text{b},\mathbf{P}_\text{b}^{'};\boldsymbol{\hat{\pi}})
  \quad \text{s.t. } \eqref{eq:bidding2}-\eqref{eq:bidding6},
  \label{eq:DFL4} \\
& \mathbf{P}_\text{b}^{'\ast}(\boldsymbol{\hat{\pi}},\mathbf{\hat{p}}),\mathbf{P}_\text{s}^{\ast} = \arg \min_{\mathbf{P}_\text{b}^{'},\mathbf{P}_\text{s}} f(\mathbf{P}_\text{b}^{\ast}(\boldsymbol{\hat{\pi}},\mathbf{\hat{p}}),\mathbf{P}_\text{b}^{'};\boldsymbol{\pi}) \label{eq:DFL5} \\
& \text{s.t. } \eqref{eq:realtime2}-\eqref{eq:realtime3}. \nonumber
\end{align}
\end{subequations}
The expectation in \eqref{eq:DFL1} is taken over the joint distribution of contextual features and ground-truth values. This formulation defines DFL under a single task setting. However, since the PLS configuration evolves dynamically with vessel arrival conditions, the following section extends this formulation to a continual learning setting on a stream of tasks.

A useful distinction in the DFL formulation is between parametric and structural task changes. Parametric changes refer to variations in exogenous realizations or arrival statistics that leave the optimization template fixed, such as different price/load samples or different arrival-time and cargo realizations under an unchanged vessel set size and constraint pattern. Structural changes refer to changes in $\boldsymbol{\xi}_d$ that modify the optimization template itself, for example through a different number of vessels $J$, different berth-service couplings, or different time-window constraints. In the latter case, the task changes not only in coefficients, but also in the dimension of the decision vector and the geometry of the feasible set $\mathcal{X}_{\mathrm{LG}}(\boldsymbol{\xi}_s,\boldsymbol{\xi}_d)$. In practice, vessel-arrival variations in PLS usually induce both effects jointly, while the present DFCL formulation is intended for this coupled setting.

\subsection{Decision-Focused Continual Learning for PLS Task Stream}

\begin{figure}[!ht]
\centering
\includegraphics[width=0.48\textwidth]{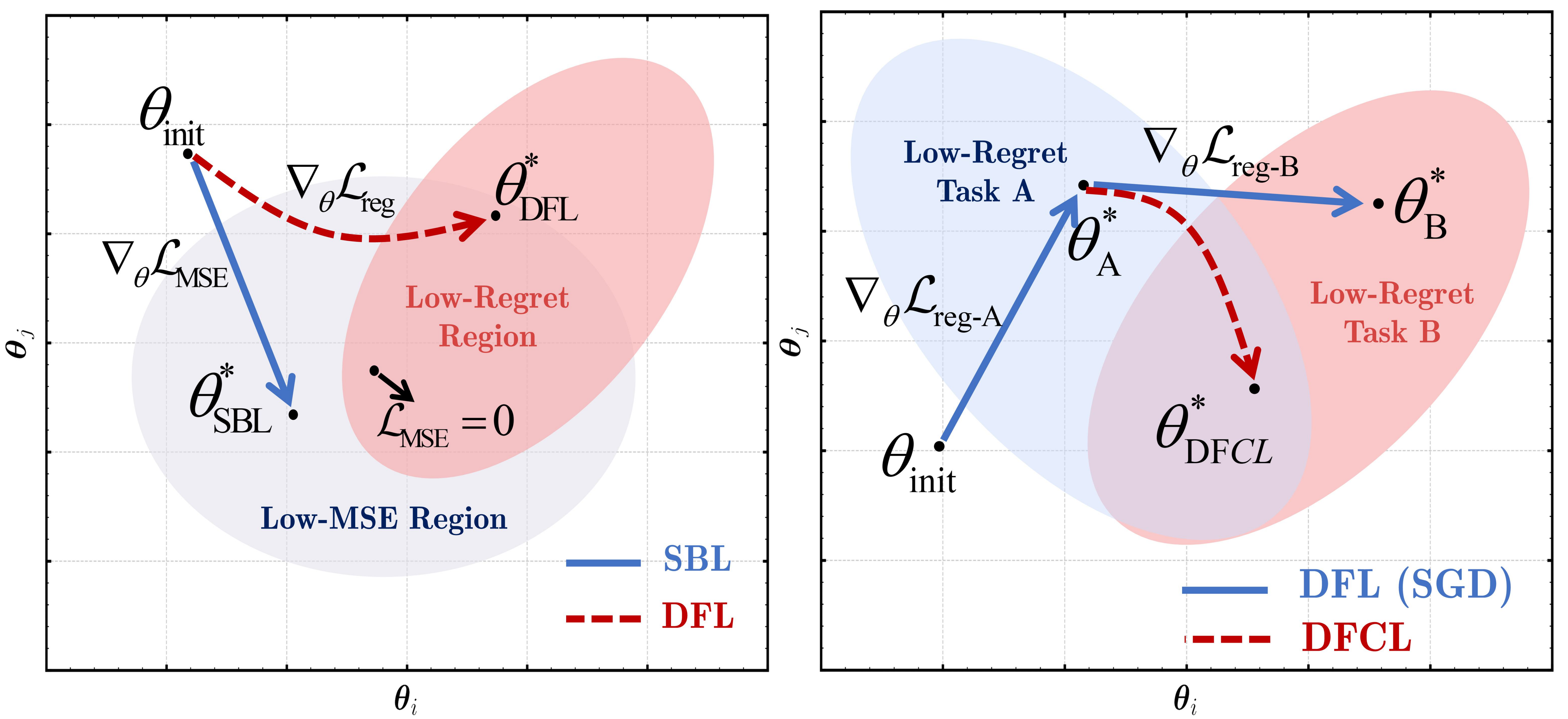}
\caption{Illustration of the loss landscapes for SBL vs. DFL (left) and DFL vs. DFCL (right). The x-axes and y-axes represent two dimensions of the parameter space, and the differently colored regions denote low-loss areas of distinct loss landscapes. The arrows indicate different learning trajectories. }
\label{fig:losslandscape_illustration}
\end{figure}

Fig.~\ref{fig:losslandscape_illustration} schematically illustrates the principle of DFCL. As shown in the left panel, SBL and DFL tasks often correspond to distinct loss landscapes, indicated by the different shaded regions in this figure. Statistical loss (e.g., MSE) typically yields symmetric cross-sections, whereas regret loss tends to be more irregular. Consequently, different loss landscapes lead to different model representations under SBL and DFL. Specifically, parameters trained under SBL, denoted as $\boldsymbol{\theta}_\text{SBL}^*$, often converge to the statistical optimum but not necessarily to a low-regret region for DFL. Similarly, the distinction between DFL and DFCL can also be explained through their loss landscapes, as shown in the right panel. Starting from $\boldsymbol{\theta}_\text{init}$, stochastic gradient descent (SGD) sequentially learns Task~A and Task~B. DFL updates the parameters to the low-regret region of Task~B, resulting in $\boldsymbol{\theta}_\text{DFL}^*$, but deviates from the low-regret region of Task~A. DFCL instead retains Task~A information and guides the parameters toward $\boldsymbol{\theta}_\text{DFCL}^*$, which lies in the overlap of low-regret regions for both tasks.

Given a task stream $\{\tau_1,\tau_2,\ldots,\tau_K\}$, the ideal objective of DFCL is to learn the forecasting model parameters $\varTheta=\{\boldsymbol{\theta}_\text{p},\boldsymbol{\theta}_\text{l}\}$ that minimize the total regret across all encountered tasks. However, since DFCL can only learn one task at a time rather than performing multi-task learning, it must minimize the regret loss of the current task while preventing significant performance degradation in previous tasks, without storing or replaying task data.

For a new task $\tau_k$, the formulation of DFCL can be described through a maximum a posteriori update,
\begin{equation}
  \label{eq:DFCL_ideal}
  \begin{aligned}
    \max_{\varTheta} \quad & \mathbb{P}(\varTheta|\mathcal{D}_{1:k}) \propto \mathbb{P}(\mathcal{D}_k|\varTheta)\, \mathbb{P}(\varTheta|\mathcal{D}_{1:k-1}),
  \end{aligned}
\end{equation}
where $\mathcal{D}_k$ is the dataset for task $\tau_k$, and $\mathcal{D}_{1:k-1}$ represents the datasets of all previous tasks. Since the decision-focused training signal is regret rather than a statistical log-likelihood, the likelihood term in \eqref{eq:DFCL_ideal} should be understood through the following pseudo-likelihood assumption.

\textbf{Assumption 1 (Regret-induced pseudo-likelihood).}
For each task $k$, the regret loss $\ell(z;\varTheta)$ induces a Gibbs-type pseudo-likelihood
\[
\tilde{\mathbb{P}}_k(z\mid\varTheta)\propto \exp\!\left(-\ell(z;\varTheta)/T\right),\quad T>0,
\]
or, equivalently for a finite dataset,
\[
\tilde{\mathbb{P}}_k(\mathcal{D}_k\mid\varTheta)
\propto
\exp\!\left(-\frac{1}{T}\sum_{z\in\mathcal{D}_k}\ell(z;\varTheta)\right).
\]
This assumption encodes the monotonic probabilistic interpretation that parameter regions yielding larger decision regret have lower pseudo-likelihood. The temperature $T$ only rescales the loss and can be absorbed into the learning rate or the regularization coefficient.

Under Assumption~1, taking the negative logarithm of the pseudo-MAP objective in \eqref{eq:DFCL_ideal} yields
\begin{equation}
  \label{eq:DFCL_ideal_log}
  \begin{aligned}
    \min_{\varTheta} \quad & -\log \mathbb{P}(\mathcal{D}_k|\varTheta) - \log \mathbb{P}(\varTheta|\mathcal{D}_{1:k-1})\\
    \iff \min_{\varTheta} \quad & \left\{\mathbb{E}\bigl[\mathcal{L}_\text{reg}(\boldsymbol{\hat{\pi}},\boldsymbol{\pi},\mathbf{\hat{p}},\mathbf{p};\varTheta)|\mathcal{D}_k\bigr] - \log \mathbb{P}(\varTheta|\mathcal{D}_{1:k-1})\right\}.
  \end{aligned}
\end{equation}
The first term in \eqref{eq:DFCL_ideal_log} corresponds to minimizing the regret loss on the current task, while the second term represents maximizing the prior knowledge distilled from previous tasks into the parameter distribution. 

Given the posterior $\mathbb{P}(\varTheta|\mathcal{D}_{1:k-1})$ after tasks $\mathcal{D}_{1:k-1}$, we apply a Laplace approximation \cite{Laplace} to approximate it as a Gaussian distribution
\begin{equation}
  \label{eq:laplace}
  \mathbb{P}(\varTheta|\mathcal{D}_{1:k-1}) \approx \mathcal{N}(\varTheta;\varTheta_{k-1}^{\ast},\mathbf{\Lambda}_{k-1}^{-1}),
\end{equation}
where $\varTheta_{k-1}^{\ast}$ is the optimal parameter obtained after learning task $\tau_{k-1}$, and $\mathbf{\Lambda}_{k-1}$ is the precision matrix defined as the Hessian of the negative log-posterior at $\varTheta_{k-1}^{\ast}$. Substituting \eqref{eq:laplace} into \eqref{eq:DFCL_ideal_log}, the negative log-posterior can be expressed as
\begin{equation}
  \label{eq:neg_log_post}
  \begin{aligned}
  -\log \mathbb{P}(\varTheta|\mathcal{D}_{1:k-1}) = \frac{1}{2}(\varTheta-\varTheta_{k-1}^{\ast})^\top \mathbf{\Lambda}_{k-1} (\varTheta-\varTheta_{k-1}^{\ast}) \\+ \frac{d}{2}\log(2\pi) - \frac{1}{2}\log|\mathbf{\Lambda}_{k-1}|,
  \end{aligned}
\end{equation}
where $d$ is the dimension of $\varTheta$. Ignoring the constant terms in \eqref{eq:neg_log_post} and substituting this approximation into \eqref{eq:DFCL_ideal_log}, the DFCL problem in $\tau_k$ can be reformulated as
\begin{equation}
  \label{eq:DFCL_final}
  \begin{aligned}
    \min_{\varTheta} \quad \{ \mathbb{E}\bigl[\mathcal{L}_\text{reg}(\boldsymbol{\hat{\pi}},\boldsymbol{\pi},\mathbf{\hat{p}},\mathbf{p};\varTheta)|\mathcal{D}_k\bigr]\\ +\frac{\lambda}{2}(\varTheta-\varTheta_{k-1}^{\ast})^\top \mathbf{\Lambda}_{k-1} (\varTheta-\varTheta_{k-1}^{\ast}) \}.
  \end{aligned}
\end{equation}
The objective of DFCL can be interpreted as minimizing the regret loss on the current task while mitigating forgetting on previous tasks through a quadratic regularization term. Here, $\lambda$ is a scalar coefficient that governs the strength of Fisher-based consolidation. Since $\mathbf{\Lambda}_{k-1}$ is estimated from empirical Fisher statistics, $\lambda$ is introduced to regulate the trade-off between adaptation to the current task and retention of decision-relevant knowledge from previous tasks.

Calculating the precision matrix $\mathbf{\Lambda}_{k-1}$ is a crucial step in implementing DFCL. Because $\mathcal{L}_{\mathrm{reg}}$ is not a literal negative log-likelihood, directly applying the classical FIM to regret would be heuristic without Assumption~1. Under the regret-induced pseudo-likelihood, we use the outer product of per-sample regret gradients as a Fisher-like positive semidefinite curvature surrogate; the detailed derivation is provided in Appendix~\ref{sec:appendixB}. Accordingly, $\mathbf{\Lambda}_{k-1}$ can be approximated by the empirical Fisher-like matrix $\widehat{\mathbf{F}}_t(\varTheta_{k-1}^{\ast})$ by sample averaging:
\begin{equation}
\label{eq:emp_fisher}
\mathbf{\Lambda}_{k-1}=\sum_{t=1}^{k-1}\beta_t\,\widehat{\mathbf{F}}_t(\varTheta_{k-1}^{\ast}),
\end{equation}
where $\beta_t$ is a scaling factor that can be interpreted as the number of samples in task $\tau_t$, and $\widehat{\mathbf{F}}_t(\varTheta)$ can be estimated by 
\begin{equation}
\label{eq:fisher_proxy}
\widehat{\mathbf{F}}_t(\varTheta)
~=~
\frac{1}{|\mathcal{D}_t|}\sum_{n=1}^{|\mathcal{D}_t|}
\nabla_{\varTheta}\mathcal{L}_{\mathrm{reg}}(z^{(t)}_n;\varTheta)\,
\nabla_{\varTheta}\mathcal{L}_{\mathrm{reg}}(z^{(t)}_n;\varTheta)^\top,
\end{equation}
where $z^{(t)}_n=(\boldsymbol{\phi}_{\mathrm{p}},\boldsymbol{\phi}_{\mathrm{l}},\boldsymbol{\pi},\mathbf{p})_{n}^{(t)}$ is the $n$-th sample. In the pseudo-likelihood interpretation, larger entries or eigenvalues of $\widehat{\mathbf{F}}_t$ indicate parameter directions to which the decision-regret landscape is more sensitive. These directions are therefore assigned stronger consolidation during the learning of new tasks to mitigate forgetting.

\textbf{Remark 1:}  Unlike naive online learning that relies solely on current-task gradients, DFCL consolidates decision-relevant knowledge from past tasks through the Fisher-like curvature surrogate induced by Assumption~1, thereby achieving continual adaptation without data replay. 

\textbf{Remark 2:} The DFCL framework is general and can be applied to various forecasting models and task configurations. The key requirement is the ability to compute the gradients of regret loss with respect to the forecasting parameters, which can be achieved through differentiable optimization techniques.

\section{Solution Method for DFCL in PLS}
\label{sec:4}
Solving the DFCL problem relies on gradient-based optimization, where ensuring end-to-end differentiability is essential to enable automatic differentiation through the computation graph. To this end, a complete end-to-end training framework is proposed, as illustrated in Fig.~\ref{fig:pipeline}, with detailed descriptions provided in the following subsections. The framework consists of three layers: the forecasting layer, the surrogate layer, and the optimization layer. In the forecasting layer, a neural-network-based model jointly processes contextual features of price and net load along with dynamic vessel features. In the surrogate layer, since existing differentiable optimization methods do not support direct differentiation through non-convex constrained problems \cite{OptNet}, a convexified surrogate model is introduced for the non-convex feasible set of the PLS scheduling problem. This surrogate predicts local variables $\mathbf{P}_\text{QC}$ and $\mathbf{V}$ based on upstream forecasts and fixes them to ensure that the remaining subproblem is convex. In the optimization layer, the PLS scheduling problem is solved based on the forecasting results and the surrogate model, and the regret loss is computed for backpropagation and calculation of the Fisher information matrix in \eqref{eq:fisher_proxy}.

\begin{figure*}[!ht]
\centering
\includegraphics[width=6in]{pipeline}
\caption{End-to-end training pipeline for DFCL in PLS. Each module represents a differentiable computational unit.
}
\label{fig:pipeline}
\end{figure*}

\subsection{Forecasting Model}
The forecasting model consists of a Transformer Encoder and a Multi-Layer Perceptron (MLP), as shown in Fig.~\ref{fig:pipeline}. Its mathematical formulation is already provided in \eqref{eq:forecasting}. Since the number of arriving vessels is variable, the Transformer Encoder is employed to process dynamic vessel-related features $\boldsymbol{\xi}_\text{d}$ to obtain a fixed-length representation, which is concatenated with static contextual features and passed to the MLP for prediction. Separate Transformer–MLP modules are designed for net load and price forecasting. For net load forecasting, the selected contextual features include historical residential load data from the past 72 hours and one week ago, one-hot encoded timestamp features, and irradiance data. For price forecasting, the chosen contextual features comprise historical electricity prices from the past 72 hours and one week ago, along with one-hot encoded timestamp features.

\subsection{Gradient Estimation for PLS Scheduling}
\subsubsection{Problem Convexification via Surrogate Model}
A critical challenge in enabling gradient-based learning for PLS lies in the non-convexity of the embedded optimization problem, which arises primarily from discrete decision variables within the feasible set $\mathcal{X}_{\text{LG}}(\boldsymbol{\xi}_s,\boldsymbol{\xi}_d)$ in \eqref{eq:bidding3}. These variables govern vessel sequencing, berth allocation, and quay-crane scheduling, leading to a combinatorial structure that prevents direct differentiation through the optimization layer. To enable end-to-end gradient back-propagation, the PLS scheduling problem must therefore be convexified while preserving its operational characteristics.

To this end, we observe that discrete logistics decisions ultimately influence the power–logistics coupling model only through the aggregated load consumption variables $(\mathbf{P}_{\text{QC}}, \mathbf{P}_{\text{s}})$ that appear in the nodal power balance constraints. Hence, the original non-convex feasible region can be relaxed by fixing $(\mathbf{P}_{\text{QC}}, \mathbf{P}_{\text{s}})$ to suitable surrogate values. Since $\mathbf{P}_{\text{s}}$ belongs to a convex set when the variable $\mathbf{V}$ is known and $\mathbf{P}_{\text{s}}$ is determined in real-time operation, we approximate $(\mathbf{P}_{\mathrm{QC}},\mathbf{V})$ using a memory-based surrogate model.

A K-Nearest Neighbors (KNN) regression is employed as the surrogate due to its non-parametric nature, which allows adaptive fitting across multiple tasks without catastrophic interference with prior knowledge. This design mitigates overfitting to offline scheduling samples, thereby yielding stable and interpretable gradient estimates. In implementation, the surrogate memory is capped by a user-specified budget $M$, so $\mathcal{M}$ stores only a bounded set of scheduling samples rather than an ever-growing history; hence, its storage complexity remains $O(M)$ with respect to the number of learned tasks. Let upstream forecasts be $\widehat{\mathbf{u}}=[\hat{\mathbf{p}},\hat{\boldsymbol{\pi}}]$, and construct a memory set as
\begin{equation}
\mathcal{M}=\Big\{\big(\mathbf{u}^{(i)},\,\mathbf{P}_{\mathrm{QC}}^{(i)},\,\mathbf{V}^{(i)}\big)\Big\}_{i=1}^{M},
\end{equation}
For a query $\widehat{\mathbf{u}}$ in forward propagation, similarities are first evaluated against all buffer samples, and the surrogate is then computed as a soft top-$K$ weighted average of their outputs, as
\begin{equation}\label{eq:softknn-y}
(\hat{\mathbf{P}}_{\mathrm{QC}}, \hat{\mathbf{V}})(\widehat{\mathbf{u}})=
\sum_{i=1}^{M} \alpha_i(\widehat{\mathbf{u}})(\mathbf{P}_{\mathrm{QC}}^{(i)}, \mathbf{V}^{(i)}),
\end{equation}
where $\alpha_i(\widehat{\mathbf{u}})$ are normalized nonnegative weights induced by a differentiable soft top-$K$ operator, so that the dominant mass is assigned to the $K$ most similar samples without hard truncation. Detailed formulations of the differentiable top-$k$ operator are provided in Appendix~\ref{sec:appendixC}.

By substituting $(\hat{\mathbf{P}}_{\text{QC}}, \hat{\mathbf{V}})$ into the original formulation, the convexified feasible region of \eqref{eq:bidding} can be expressed as
\begin{equation}
\mathcal{X}_{\text{convex}}(\boldsymbol{\xi}_s,\hat{\mathbf{V}},\hat{\mathbf{P}}_{\text{QC}}) =
\mathcal{X}_{\text{ES}}(\boldsymbol{\xi}_s) \;\cap\;
\bigl\{ \mathbf{P}_{\text{s}} \;\big|\; \eqref{eq:bidding5}, \eqref{eq:bidding6},\eqref{eq:shore_power} \bigr\}.
\end{equation}

\subsubsection{Differentiation of the Regret Loss}
The remaining challenge is to compute the gradient of the regret loss $\mathcal{L}_{\mathrm{reg}}$ with respect to the forecasting parameters $\varTheta$. Let the day-ahead decision be
\(\mathbf{x}_{D}^{\ast}(\boldsymbol{\eta})
=
\bigl[\mathbf{P}_\text{b}^{\ast},\mathbf{P}_\text{ES.dch}^{\ast},\mathbf{P}_\text{ES.ch}^{\ast},\mathbf{P}_\text{QC}^{\ast},\mathbf{V}^{\ast}\bigr]\)
from \eqref{eq:DFL4} and the real-time decision is
\(\mathbf{x}_{R}^{\ast}(\boldsymbol{\eta},\mathbf{x}_{D}^{\ast})
=
\bigl[\mathbf{P}_\text{b}^{'\ast},\mathbf{P}_\text{s}^{\ast}\bigr]\)
from \eqref{eq:DFL5}. In the day-ahead problem, the inputs related to $\varTheta$ include the predicted electricity prices, net load, and the outputs of the surrogate model $(\hat{\mathbf{P}}_{\text{QC}}, \hat{\mathbf{V}})$, collectively denoted as $\boldsymbol{\eta}=[\boldsymbol{\hat{\pi}},\mathbf{\hat{p}},\hat{\mathbf{P}}_{\text{QC}}, \hat{\mathbf{V}}]$. The real-time problem takes the day-ahead decision $\mathbf{x}_{D}^{\ast}(\boldsymbol{\eta})$ as input. According to the chain rule, the gradient w.r.t.\ any predicted quantity \(\boldsymbol{\eta}\) is
\begin{align}
\frac{d\mathcal{L}_{\mathrm{reg}}}{d\boldsymbol{\eta}}
&=
\frac{\partial\mathcal{L}_{\mathrm{reg}}}{\partial\boldsymbol{\eta}}
\;+\;
\frac{\partial\mathcal{L}_{\mathrm{reg}}}{\partial\mathbf{x}_{D}^{\ast}}
\frac{d\mathbf{x}_{D}^{\ast}}{d\boldsymbol{\eta}}
\;+\;
\frac{\partial\mathcal{L}_{\mathrm{reg}}}{\partial\mathbf{x}_{R}^{\ast}}
\frac{d\mathbf{x}_{R}^{\ast}}{d\boldsymbol{\eta}}.
\label{eq:chainrule-top}
\end{align}
Since $\mathcal{L}_{\mathrm{reg}}$ is an explicit function, the partial derivatives $\partial\mathcal{L}_{\mathrm{reg}}/\partial\mathbf{x}_{D}^{\ast}$ and $\partial\mathcal{L}_{\mathrm{reg}}/\partial\mathbf{x}_{R}^{\ast}$ can be computed directly. The remaining challenge is to compute the implicit gradients $\frac{d\mathbf{x}_{D}^{\ast}}{d\boldsymbol{\eta}}$ and $\frac{d\mathbf{x}_{R}^{\ast}}{d\boldsymbol{\eta}}$. 
The following describes how to implicitly compute the gradients of the optimal solution with respect to input parameters from the KKT conditions.

Both the day-ahead problem \eqref{eq:DFL4} and the real-time problem \eqref{eq:DFL5} admit a standard conic-QP form
\begin{subequations}
\label{eq:conic-form}
\begin{align}
\min_{\mathbf{x},\,\mathbf{s}}~
& \tfrac{1}{2}\mathbf{x}^{\top}\mathbf{P}\mathbf{x} + \mathbf{q}(\boldsymbol{\eta})^{\top}\mathbf{x} \\
\text{s.t.}~
& \mathbf{A}(\boldsymbol{\eta})\mathbf{x} + \mathbf{s} = \mathbf{b}(\boldsymbol{\eta}),~~ \mathbf{s} \in \mathcal{K}, 
\end{align}
\end{subequations}
where the prediction vector enters the linear cost \(\mathbf{q}(\boldsymbol{\eta})\) and the linear maps \(\mathbf{A}(\boldsymbol{\eta})\), \(\mathbf{b}(\boldsymbol{\eta})\).
In our setting, \(\mathbf{P}=\mathbf{0}\) and \(\mathcal{K}=\mathbb{R}_{+}^{m}\) (purely linear constraints), i.e., \eqref{eq:conic-form} reduces to an LP. When needed to ensure differentiability, we consider a standard Tikhonov regularization \(\epsilon\|\mathbf{x}\|^{2}/2\) (i.e., \(\mathbf{P}=\epsilon\mathbf{I}\), \(\epsilon>0\)), which leaves the optimizer unchanged in the limit \(\epsilon\downarrow 0\) but yields a well-conditioned KKT system.

Let \(\boldsymbol{\lambda}\) be the dual variable of the conic constraint and \(\mathbf{s}\in\mathcal{K}\) the slack.
The KKT conditions at a primal-dual solution \((\mathbf{x}^{\ast},\mathbf{s}^{\ast},\boldsymbol{\lambda}^{\ast})\) are
\begin{subequations}
\label{eq:KKT}
\begin{align}
\text{(stationarity)}\quad
& \mathbf{P}\mathbf{x}^{\ast} + \mathbf{q}(\boldsymbol{\eta}) + \mathbf{A}(\boldsymbol{\eta})^{\top}\boldsymbol{\lambda}^{\ast} = \mathbf{0}, \label{eq:KKT-sta}\\
\text{(primal feasibility)}\quad
& \mathbf{A}(\boldsymbol{\eta})\mathbf{x}^{\ast} + \mathbf{s}^{\ast} - \mathbf{b}(\boldsymbol{\eta}) = \mathbf{0}, \label{eq:KKT-pri}\\
\text{(dual feasibility)}\quad
& \mathbf{s}^{\ast}\in\mathcal{K},~~ \boldsymbol{\lambda}^{\ast}\in\mathcal{K}^{\ast}, \label{eq:KKT-feas}\\
\text{(complementarity)}\quad
& \langle \mathbf{s}^{\ast}, \boldsymbol{\lambda}^{\ast}\rangle = 0. \label{eq:KKT-comp}
\end{align}
\end{subequations}
For \(\mathcal{K}=\mathbb{R}_{+}^{m}\), \eqref{eq:KKT-feas}–\eqref{eq:KKT-comp} reduce to \(\mathbf{s}^{\ast}\ge 0\), \(\boldsymbol{\lambda}^{\ast}\ge 0\), and componentwise to \(s_{i}^{\ast}\lambda_{i}^{\ast}=0\). Denote \(\mathbf{S}:=\mathrm{diag}(\mathbf{s}^{\ast})\) and \(\mathbf{D}:=\mathrm{diag}(\boldsymbol{\lambda}^{\ast})\). Define the residual mapping \(\mathbf{H}(\mathbf{z},\boldsymbol{\eta})=\mathbf{0}\) with \(\mathbf{z}:=(\mathbf{x},\mathbf{s},\boldsymbol{\lambda})\) by stacking \eqref{eq:KKT-sta}–\eqref{eq:KKT-pri} and a differentiable surrogate for complementarity (e.g., \(\mathbf{S}\boldsymbol{\lambda}=\mathbf{0}\) on the active set). Linearizing \(\mathbf{H}\) around \((\mathbf{z}^{\ast},\boldsymbol{\eta})\) gives
\begin{align}
\underbrace{\mathbf{K}}_{:=~\partial_{\mathbf{z}}\mathbf{H}(\mathbf{z}^{\ast},\boldsymbol{\eta})}\, d\mathbf{z}
~=~
-\, \underbrace{\partial_{\boldsymbol{\eta}}\mathbf{H}(\mathbf{z}^{\ast},\boldsymbol{\eta})}_{=:~\mathbf{J}_{\eta}}\, d\boldsymbol{\eta}.
\label{eq:lin-sys}
\end{align}
Under standard regularity, \(\mathbf{K}\) is nonsingular on the pertinent subspace and hence
\begin{align}
\frac{d\mathbf{z}^{\ast}}{d\boldsymbol{\eta}} 
~=~ -\, \mathbf{K}^{-1}\mathbf{J}_{\eta}.
\label{eq:sensitivity}
\end{align}
The solution of \eqref{eq:sensitivity} contains the sensitivities of all primal and dual variables. We only need the primal part \(\frac{d\mathbf{x}^{\ast}}{d\boldsymbol{\eta}}\) for the chain rule \eqref{eq:chainrule-top}. For the LP/QP with a non-negative-orthant cone, the Jacobian \(\mathbf{K}\) takes the explicit block form
\begin{align}
\mathbf{K}
=
\begin{bmatrix}
\mathbf{P} & \mathbf{0} & \mathbf{A}^{\top} \\
\mathbf{A} & \mathbf{I} & \mathbf{0} \\
\mathbf{0} & \mathbf{D} & \mathbf{S}
\end{bmatrix},
\qquad
\mathbf{J}_{\eta}
=
\begin{bmatrix}
\partial_{\boldsymbol{\eta}}\mathbf{q} + (\partial_{\boldsymbol{\eta}}\mathbf{A})^{\top}\boldsymbol{\lambda}^{\ast} \\
\partial_{\boldsymbol{\eta}}\mathbf{A}\,\mathbf{x}^{\ast} - \partial_{\boldsymbol{\eta}}\mathbf{b} \\
\mathbf{0}
\end{bmatrix}.
\label{eq:KKT-Jacs}
\end{align}
Here, \(\mathbf{A},\mathbf{b},\mathbf{q}\) are evaluated at \((\mathbf{x}^{\ast},\boldsymbol{\eta})\) and we suppress the \(\boldsymbol{\eta}\)-dependence for readability. 
The derivatives \(\partial_{\boldsymbol{\eta}}\mathbf{A}\), \(\partial_{\boldsymbol{\eta}}\mathbf{b}\), and \(\partial_{\boldsymbol{\eta}}\mathbf{q}\) are straightforward to compute from the problem data. The sensitivity \eqref{eq:sensitivity} can be calculated by solving the linear system \eqref{eq:lin-sys} using the block structure of \(\mathbf{K}\) for efficiency.

\section{Case Study}
\label{sec:5}

\subsection{Experimental Setup}

\subsubsection{PLS Data Configuration}
The case study is configured as a representative seaport PLS benchmark inspired by the operating scale and port type of Jurong Port in Singapore. The port is equipped with ten quay cranes (70 TEU/h, 0.32 MW each), a total berth length of 800 m, and a 5 MW/3 h battery ESS (efficiency 0.9). For day-ahead bidding, imbalance penalty factors are set to $\rho^{+}=1.8$ for upward deviation and $\rho^{-}=0.5$ for downward deviation. 

\subsubsection{Forecasting Datasets}
Due to the lack of publicly available co-located port-level datasets that simultaneously provide vessel arrivals, PV generation, load, and market prices, we construct a reproducible benchmark by combining representative vessel-arrival and shore-power task scenarios with publicly available PV/price/load time series. The PV generation dataset was compiled from historical measurements in PES Region 10 of East England, spanning September 2020 to January 2024, and supplemented with meteorological variables from the DWD Center. The corresponding electricity price dataset was collected from the UK day-ahead market over the same horizon. Both datasets are publicly accessible in \cite{HEFTCom}. The residential load dataset was obtained from Zone 1 of the GEFCom2014 competition \cite{GEFCom2014}.


\subsubsection{Task Settings}
Six different task scenarios are constructed to simulate online continual learning in PLS. 
Since vessel arrival patterns are independent of market price and load data, to isolate the impact of uneven temporal distributions in the exogenous series, each task $k$ shares the same datasets for net load and electricity prices, denoted as $\mathcal{D}_\text{l}^{(k)}=\{(\boldsymbol{\phi}_\text{l}^{(n)},\mathbf{p}^{(n)})\}_{n=1}^{N} \cup \boldsymbol{\xi}_\text{d}^{(k)}$ and $\mathcal{D}_\text{p}^{(k)}=\{(\boldsymbol{\phi}_\text{p}^{(n)},\boldsymbol{\pi}^{(n)})\}_{n=1}^{N} \cup \boldsymbol{\xi}_\text{d}^{(k)}$, while differing only in vessel-arrival conditions. This design ensures that performance differences across tasks are primarily attributable to vessel-arrival patterns and associated scheduling-structure changes, rather than to coincidental changes in price, weather, or load samples. Consequently, although the benchmark does not eliminate all realism gaps induced by cross-region data assembly, these gaps affect all methods under the same protocol and do not confound the central comparison on cross-task adaptation. The detailed vessel arrival information for each task is summarized in Table~\ref{tab:summary_statistics} in Appendix~\ref{sec:appendixD}, including arrival times, latest departure times, cargo volumes, and the required minimum and maximum numbers of quay cranes. To ensure a robust evaluation of generalization performance, each task scenario includes 402 days of training data and 400 days of testing data.

\subsection{Results of DFL vs. SBL}
\label{sec:Q1}
To evaluate the potential of DFL in reducing operational costs, we compare DFL with several baseline methods on Task~1 as a representative case. The baselines include the SBL approach, the DFL-\cite{DFL-ED3} method that uses warm-start and heuristic strategies for gradient-free optimization of the forecasting model, and the DFL-\cite{DFL-surrogate} method that employs a neural network surrogate to enable differentiation through the embedded optimizer. All DFL methods use the same neural network architecture, hyperparameters, and early-stopping settings on the Task~1 training set. The performance metrics on the test set are summarized in Table~\ref{tab:performance_comparison}.

\begin{table}[!ht]
\centering
\begin{threeparttable}
\caption{Performance comparison of different methods.}
\label{tab:performance_comparison}
\begin{tabular}{
    L{1.2cm}  
    C{0.7cm}  
    C{0.7cm}  
    C{0.9cm}  
    C{0.9cm}  
    C{0.8cm}  
    C{0.8cm}  
}
\hline\hline
\textbf{Method} & \textbf{MAE (Load)} & \textbf{MAE (Price)} & \textbf{Cost (\$)} & \textbf{Regret (\$)} & $\boldsymbol{\Delta}$ (\%) & \textbf{Regret (\%)} \\
\midrule
SBL               & 2.08 & 13.83 & 71451.64 & 3305.85 & /    & 4.85 \\
DFL-\cite{DFL-ED3} & 2.17 & 13.92 & 71347.43 & 3201.65 & 3.15 & 4.70 \\
DFL-\cite{DFL-surrogate}  & 2.23 & 27.36 & 71447.11 & 3301.32 & 0.14 & 4.84 \\
DFL*              & 2.25 & 14.22 & 71252.49 & 3106.71 & 6.02 & 4.56 \\
\hline\hline
\end{tabular}
\begin{tablenotes}[flushleft]
\footnotesize
\item[*] denotes the proposed method in this paper.
\end{tablenotes}
\end{threeparttable}
\end{table}

From Table~\ref{tab:performance_comparison}, all variants of DFL reduce the total operating cost relative to SBL, with the proposed method achieving the lowest cost. This improvement is notable given the small differences in net load and price MAE across methods. The regret values further illustrate the performance gap, with the proposed DFL method exhibiting the lowest regret of \$3,106.71 (4.56\%), compared to \$3,305.85 (4.85\%) for SBL. The gradient-free DFL-\cite{DFL-ED3} delivers limited gains due to its search strategy, while the surrogate-based DFL-\cite{DFL-surrogate} suffers from overfitting of gradient estimates, degrading forecast quality and suppressing regret reduction. By contrast, the proposed model-free surrogate provides stable decision gradients even under non-convex and discrete scheduling, enabling more effective parameter updates. Since all methods are evaluated on the same mixed-source benchmark under identical task definitions, these gains should be interpreted as evidence of relative algorithmic advantage under matched protocols, rather than as a claim that the assembled dataset fully reproduces Singapore-specific climate or market conditions.

\begin{figure*}[!ht]
\centering
{
\includegraphics[width=0.32\textwidth]{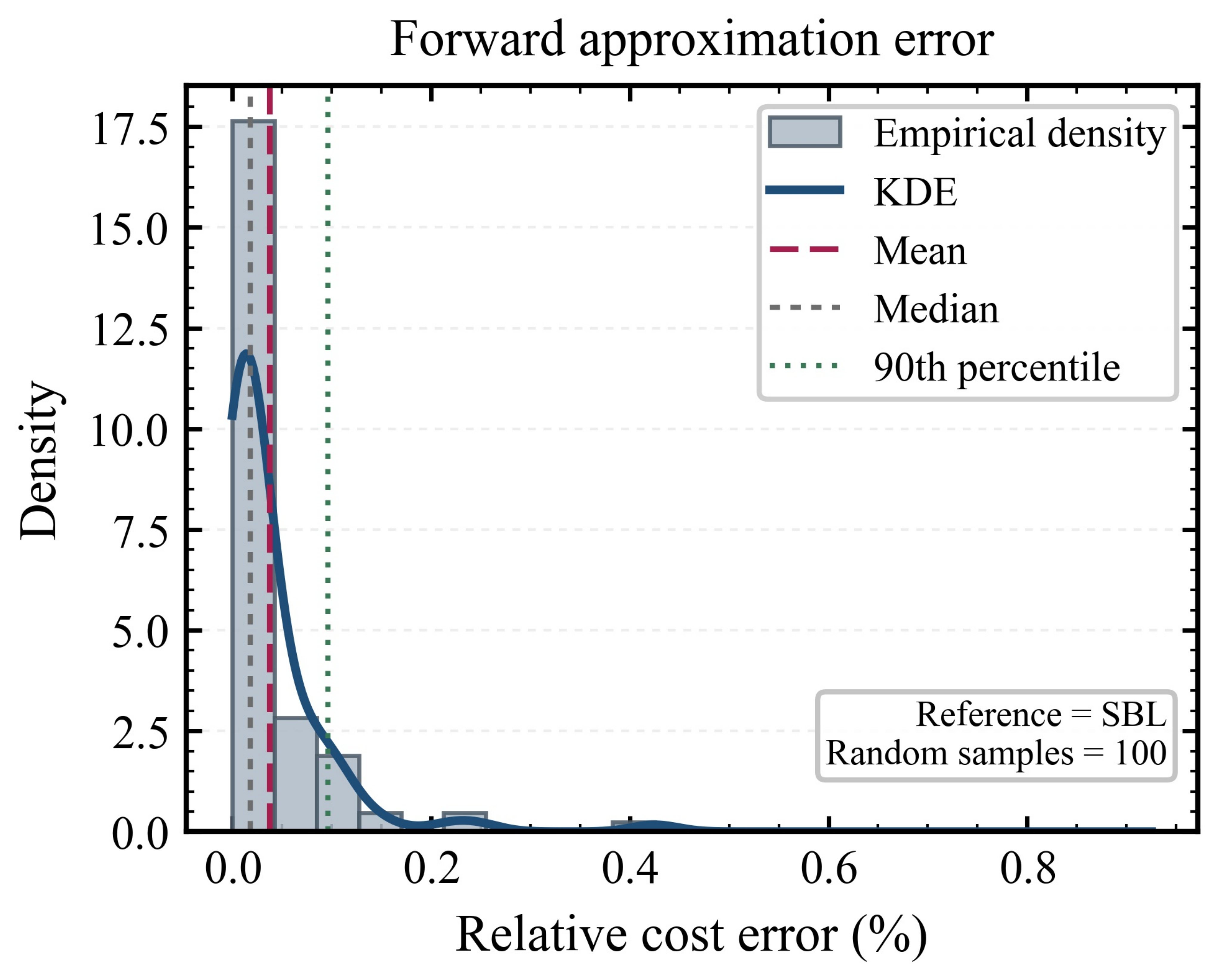}
\label{fig:surrogate_validation_forward}
}
\hfill
{
\includegraphics[width=0.6\textwidth]{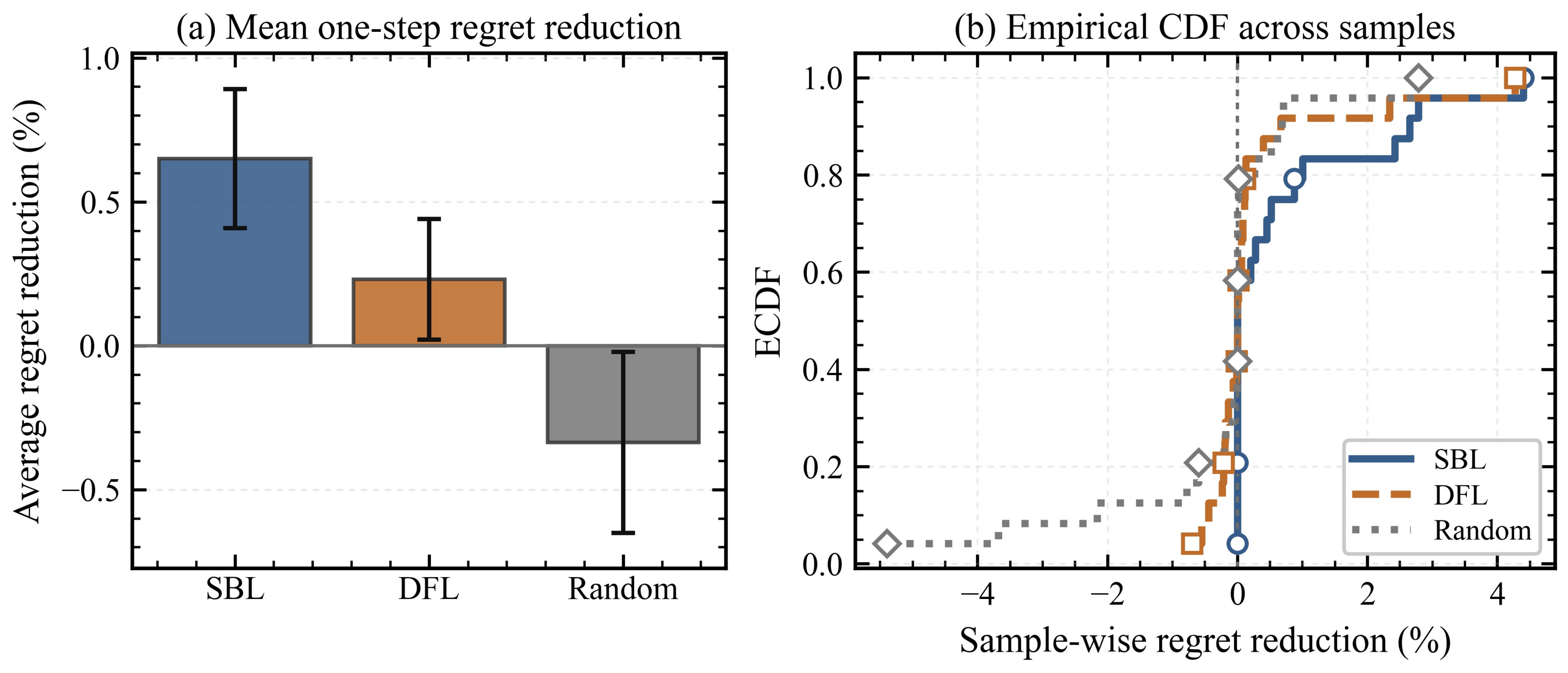}
\label{fig:surrogate_validation_direction}
}
\caption{Empirical validation of the differentiable surrogate used in DFCL training. Panel (a) plots the distribution of the relative forward cost error $\epsilon_{\mathrm{fwd}}$ over 100 random instances; Panel (b) compares three update directions using the one-step regret reduction $\Delta_{\mathrm{reg}}$: the left bar chart reports the sample mean with error bars, and the right ECDF reports the cumulative fraction of instances whose sample-wise regret reduction is no greater than the horizontal-axis value.}
\label{fig:surrogate_validation}
\end{figure*}

\subsection{Validation of the Differentiable Surrogate}
\label{sec:surrogate_validation}
To further examine whether the proposed surrogate is sufficiently faithful for back-propagation, we validate it from two complementary aspects: the local forward effect induced by a surrogate step and the directional consistency of the induced gradient. The validation is conducted on random test instances drawn from all six tasks using the converged SBL predictor as a common starting point. For each sampled instance, we first solve the original mixed-integer day-ahead and real-time scheduling problems to obtain the realized operating cost $C(\hat{\boldsymbol{\pi}},\hat{\mathbf{p}})$ under the predicted price and net load. We then compute the surrogate gradient with respect to the forecast vector $\hat{\mathbf{u}}=[\hat{\boldsymbol{\pi}}^{\top},\hat{\mathbf{p}}^{\top}]^{\top}$ and normalize it into a unit direction $\mathbf{d}_{\mathrm{DFL}}=[\mathbf{d}_{\mathrm{DFL}}^{\pi\top},\mathbf{d}_{\mathrm{DFL}}^{p\top}]^{\top}$. A fixed step of size $\eta=5$ is then applied to form the perturbed forecasts $\tilde{\boldsymbol{\pi}}=\hat{\boldsymbol{\pi}}+\eta\mathbf{d}_{\mathrm{DFL}}^{\pi}$ and $\tilde{\mathbf{p}}=\hat{\mathbf{p}}+\eta\mathbf{d}_{\mathrm{DFL}}^{p}$.

For Fig.~\ref{fig:surrogate_validation}(a), the forward metric is defined as
\begin{equation}
\epsilon_{\mathrm{fwd}}
=
\frac{\left|C(\tilde{\boldsymbol{\pi}},\tilde{\mathbf{p}})-C(\hat{\boldsymbol{\pi}},\hat{\mathbf{p}})\right|}
{\left|C(\hat{\boldsymbol{\pi}},\hat{\mathbf{p}})\right|}
\times 100\%,
\end{equation}
which measures the relative change in the realized cost of the original scheduling pipeline after one normalized surrogate step in the forecast space. The histogram in Fig.~\ref{fig:surrogate_validation}(a) summarizes this quantity over 100 randomly sampled instances. The resulting variation remains consistently small, with a mean of 0.037\%, a median of 0.018\%, and a 90th percentile of 0.096\%. This indicates that the local convexification used during back-propagation does not introduce large forward perturbations in the neighborhood relevant for parameter updates.

For Fig.~\ref{fig:surrogate_validation}(b), we further test whether the surrogate gradient provides a useful descent direction for the original scheduling problem. Starting from the same SBL prediction, we compare three unit directions in the forecast space under the same step size $\eta$: i) the \emph{SBL direction} $\mathbf{d}_{\mathrm{SBL}}\propto [\,(\boldsymbol{\pi}-\hat{\boldsymbol{\pi}})^{\top},(\mathbf{p}-\hat{\mathbf{p}})^{\top}\,]^{\top}$, ii) the \emph{DFL direction} $\mathbf{d}_{\mathrm{DFL}}$ returned by the surrogate gradient, and iii) a random unit vector. For each direction $\mathbf{d}=[\mathbf{d}^{\pi\top},\mathbf{d}^{p\top}]^{\top}$, the perturbed forecasts $(\hat{\boldsymbol{\pi}}+\eta\mathbf{d}^{\pi},\hat{\mathbf{p}}+\eta\mathbf{d}^{p})$ are fed back into the original mixed-integer scheduler, and the one-step regret reduction is computed as
\begin{equation}
\Delta_{\mathrm{reg}}(\mathbf{d})
=
\frac{\mathcal{L}_{\mathrm{reg}}(\hat{\boldsymbol{\pi}},\hat{\mathbf{p}})
-\mathcal{L}_{\mathrm{reg}}(\hat{\boldsymbol{\pi}}+\eta\mathbf{d}^{\pi},\hat{\mathbf{p}}+\eta\mathbf{d}^{p})}
{\max\left(\left|\mathcal{L}_{\mathrm{reg}}(\hat{\boldsymbol{\pi}},\hat{\mathbf{p}})\right|,10^{-6}\right)}
\times 100\%.
\end{equation}
We visualize $\Delta_{\mathrm{reg}}(\mathbf{d})$ over 100 randomly sampled instances in two complementary ways. The bar chart reports the sample mean for each direction, with error bars indicating the cross-instance variability. The ECDF (empirical cumulative distribution function) is defined so that the vertical value at any horizontal coordinate $x$ gives the fraction of instances whose sample-wise regret reduction is no greater than $x$. Under this interpretation, a curve shifted further to the right indicates larger regret reductions on a larger portion of samples, whereas the ECDF value at $x=0$ gives the fraction of non-improving cases and $1-\mathrm{ECDF}(0)$ gives the fraction of improving cases. The SBL direction yields the largest average one-step regret reduction and the most favorable distribution. The surrogate-induced DFL direction still dominates the random baseline in both the mean plot and the ECDF, indicating more frequent and larger positive reductions. In particular, the DFL direction achieves regret improvement on 70.3\% of the sampled instances, compared with 29.2\% for the random direction. These results suggest that, although the surrogate is only used locally during training, the gradients it produces remain directionally informative for the true non-convex decision problem.

\vspace{-0.5em}
\subsection{Results of DFL in Continually Arriving Tasks}
\label{sec:Q2}
To evaluate the performance degradation of a single DFL model under task evolution, we first train the forecasting model using the proposed DFL method on Task~1 for 200 epochs and then directly deploy it on the test sets of Tasks~2 to Task~6 without retraining. Fig.~\ref{fig:performance_degradation} presents heat maps of the loss distribution for all test samples across the different tasks. The performance gap $\mathrm{Gap}(\%)$ is defined as
\begin{align}
\mathrm{Gap}_{i}(\%)
=
\frac{\mathbb{E}_{\mathcal{D}_i}(\mathcal{L}_\text{reg}(\varTheta^{\ast}_1)) - \mathbb{E}_{\mathcal{D}_i}(\mathcal{L}_\text{reg}(\varTheta^{\ast}_i))}
{\mathbb{E}_{\mathcal{D}_i}(\mathcal{L}_\text{reg}(\varTheta^{\ast}_i))} \times 100\%,
\label{eq:gap}
\end{align}
where $\mathbb{E}_{\mathcal{D}_i}(\mathcal{L}_\text{reg}(\varTheta^{\ast}_1))$ denotes the regret of the model trained on Task~1 when evaluated on the test set of Task~$i$, and $\mathbb{E}_{\mathcal{D}_i}(\mathcal{L}_\text{reg}(\varTheta^{\ast}_i))$ is the regret of a model trained specifically on Task~$i$.
\begin{figure}[!ht]
\centering
\includegraphics[width=0.5\textwidth]{loss_landscape}
\caption{Loss heat maps of a DFL model trained on Task~1 and evaluated on Tasks~1 to 6. The x-axis represents the load forecast error (actual minus predicted), while the y-axis represents the price forecast error. The color intensity indicates the regret value, with colder colors representing higher regret.
}
\label{fig:performance_degradation}
\end{figure}

Two empirical regularities are observed. First, regret tends to be lower near the center of each heat map, yet many off-center points also exhibit low regret. This indicates that operating cost is not determined solely by forecast error magnitude, which helps explain why DFL can outperform SBL. Second, the loss landscapes differ across tasks, which means that optimal performance on one task does not guarantee optimality on others. A single DFL model trained on Task~1 shows performance degradation ranging from 2.20\% to 8.67\% on Tasks~2 to 6, as measured by $\mathrm{Gap}(\%)$.

\subsection{Performance Evaluation of DFCL}
\label{sec:Q3}
The aforementioned performance degradation highlights the need for continual learning to adapt to evolving tasks. We evaluate the proposed DFCL method against four baselines: (i) SBL, which retrains the forecasting models from scratch for each task; (ii) DFL (Adam), which continually fine-tunes the DFL model with the Adam optimizer but without any forgetting mitigation; (iii) DFCL (Adam-F), a naïve continual variant that freezes all layers except the last during fine-tuning to preserve prior knowledge; and (iv) the proposed DFCL (EWC). All methods are trained sequentially on Tasks~1–6 under the same neural network architecture and hyperparameters. For DFCL (EWC), the regularization coefficient in \eqref{eq:DFCL_final} is set to $\lambda=3\times10^{-3}$ unless otherwise stated. After completing Task~$k$, we evaluate on the test sets of Tasks~1 through $k$ and report the average MAE of net load and price forecasts, the cumulative regret, the regret ratio (RR), and a forgetting measure (FM) that quantifies performance loss on earlier tasks after learning later ones. The regret ratio is defined as
\begin{align}
\mathrm{RR}_{k}(\%)=\frac{\sum_{i=1}^{k}\,\mathbb{E}_{\mathcal{D}_i}\!\left(\mathcal{L}_\text{reg}(\varTheta^{\ast}_k)\right)}
{\sum_{i=1}^{k}\,\mathbb{E}_{\mathcal{D}_i}\!\left(f(\cdot;\varTheta^{\ast}_k)\right)
-\sum_{i=1}^{k}\,\mathbb{E}_{\mathcal{D}_i}\!\left(\mathcal{L}_\text{reg}(\varTheta^{\ast}_k)\right)},
\label{eq:rr_def}
\end{align}
that is, $\mathrm{RR}_{k}$ measures the ratio of cumulative regret to the cumulative perfect-foresight cost over the first $k$ tasks. FM evaluates how much performance on earlier tasks deteriorates after learning new ones. For the $k$-th task, FM is defined as
\begin{equation}
\begin{aligned}
FM_k=\frac{1}{k-1}\sum_{j=1}^{k-1}
\Big[&\mathbb{E}_{\mathcal{D}_j}\!\left(\mathcal{L}_\text{reg}(\varTheta^{\ast}_k)\right) \\
&-\min_{\ell\in\{j,\ldots,k-1\}}\mathbb{E}_{\mathcal{D}_j}\!\left(\mathcal{L}_\text{reg}(\varTheta^{\ast}_{\ell})\right)\Big],
\end{aligned}
\end{equation}
that is, the average degradation of the current regret relative to the best historical regret previously achieved on each earlier task.

After continually learning the first $k$ tasks ($k=1,\dots,6$), the average performance on all previously learned tasks is summarized in Table~\ref{tab:cl_case_stacked_widths}. The regret trajectories for Tasks~1–5 across sequential training from Task~1 to Task~6, along with the overall average regret trend, are shown in Fig.~\ref{fig:regret_trend}.
Fig.~\ref{fig:regret_trend} reports the mean regret trajectories over five runs for each method, and the shaded bands indicate $\pm 1$ standard deviation.
\begin{figure*}[!ht]
\centering
\includegraphics[width=1\textwidth]{DFCL_performance_23}
\caption{Evolution of task-wise regret on Tasks~1--5 and the average regret over learned tasks during sequential training from Task~1 to Task~6. Curves report the mean over five runs, and the shaded bands indicate $\pm 1$ standard deviation.}
\label{fig:regret_trend}
\end{figure*}

From Table~\ref{tab:cl_case_stacked_widths} and Fig.~\ref{fig:regret_trend}, the proposed DFCL (EWC) achieves the lowest average regret and regret ratio after training each task and shows the most stable trend over the task streams. By contrast, DFL (Adam) exhibits pronounced forgetting during continual learning. Relative to DFL (Adam), DFCL (EWC) reduces cumulative regret by up to 2.5\% and the forgetting measure by up to 27.7\% after completing 6 tasks. This result indicates that the EWC regularization improves cross-task generalization. DFCL (Adam\textendash F) offers slight gains when consecutive tasks have similar ground truth (Tasks~5 and~6), but its forgetting measure is marginally higher on Tasks~2--4, suggesting that freezing layers alone is insufficient to prevent forgetting. Another notable finding is that SBL records the highest regret but the lowest forecast errors across all tasks, consistent with its objective of minimizing statistical error rather than decision loss. This also suggests that DFL confers an inherent degree of cross-task generalization, as illustrated by the overlapping low-regret regions in the schematic loss landscapes of Fig.~\ref{fig:losslandscape_illustration}. Importantly, the continual-learning comparison is driven by changes in vessel-arrival configurations while holding the exogenous benchmark time series fixed across tasks. The observed ranking therefore reflects robustness to task-structure shifts, which is the main claim of this paper, rather than robustness to geographic transfer across different electricity markets.

Since continual learning performance may depend on task order, we further examine the robustness of DFCL under six balanced task sequences. Let $\pi^{(m)}=(\pi^{(m)}_1,\ldots,\pi^{(m)}_6)$ denote the $m$-th sequence and let $\mathcal{T}=\{1,\ldots,6\}$ denote the task set. Besides the canonical order Task~1$\rightarrow$Task~2$\rightarrow\cdots\rightarrow$Task~6, we construct five additional sequences such that
\begin{equation}
\begin{aligned}
\{\pi^{(m)}_p\}_{m=1}^{6} &= \mathcal{T}, \quad \forall p\in\mathcal{T}, \\
\{(\pi^{(m)}_p,\pi^{(m)}_{p+1})\}_{m=1,p=1}^{6,5} &= \{(i,j)\in\mathcal{T}^2:i\neq j\}.
\end{aligned}
\end{equation}
That is, the task sequences are selected so that each task appears once at each position and each ordered pair of consecutive tasks is covered once. This design reduces order bias and provides a fairer test of robustness. The resulting RR and FM after learning all six tasks are shown in Fig.~\ref{fig:task_sequence_rr_fm}.
\begin{figure}[!ht]
\centering
\includegraphics[width=0.5\textwidth]{task_sequence_rr_fm}
\caption{Average RR and FM after learning all six tasks under six balanced task sequences that satisfy positional balance and adjacent-transition balance. Each marker style denotes one task sequence, and each color denotes one method. }
\label{fig:task_sequence_rr_fm}
\end{figure}

Fig.~\ref{fig:task_sequence_rr_fm} yields two main observations. First, all methods exhibit noticeable dispersion across sequences, confirming that task order affects the absolute continual-learning performance. Second, DFCL (EWC) dominates the other continual baselines on every tested sequence: for each marker shape, its RR--FM point lies below and to the left of the corresponding DFCL (Adam\textendash F) and DFL (Adam) points. Moreover, the DFCL (EWC) points form the tightest cluster, indicating lower sensitivity to task-order perturbations. These results show that the benefit of the proposed EWC-based regularization is not tied to a favorable ordering; rather, it consistently preserves decision-effective knowledge under varied task-transition patterns.

\subsection{Sensitivity to Imperfect Real-Time Forecasts}
The limited-recourse formulation in the PLS pipeline evaluates day-ahead forecasts using near-real-time updates as proxies for realized intraday conditions. To examine whether the comparison between SBL and DFL is unduly dependent on this approximation, we conduct a post-hoc sensitivity analysis in which the real-time net-load updates are perturbed, while the trained forecasting models and their induced day-ahead schedules are kept fixed. Specifically, for each method, each task in Tasks~1--6, and each selected test day, we first solve the original day-ahead scheduling problem using the forecasts produced by the learned model, thereby fixing the committed bid, ESS schedule, and power/logistics decisions. We then replace the true real-time net load $\mathbf{p}^{\mathrm{rt}}$ with a noisy update
\begin{equation}
\tilde{\mathbf{p}}^{\mathrm{rt}}
=
\mathbf{p}^{\mathrm{rt}}+\eta \sigma_{p}\boldsymbol{\epsilon},
\qquad
\boldsymbol{\epsilon}\sim\mathcal{N}(\mathbf{0},\mathbf{I}),
\end{equation}
where $\sigma_{p}$ is the empirical standard deviation of the task-specific test net load and $\eta$ is the real-time noise ratio. We vary $\eta$ from 0 to 0.6, evaluate 48 uniformly spaced test days per task, and repeat each noise level over seven random seeds.

The recourse problem is solved with $\tilde{\mathbf{p}}^{\mathrm{rt}}$ as the available real-time information, but the resulting operating cost and regret are evaluated ex post using the true net load. For each seed and each noise level, the regret is averaged over the sampled test days of all six tasks. We then report the \emph{DFL regret advantage over SBL},
\begin{equation}
\mathcal{A}(\eta)
=
\overline{\mathcal{L}}_{\mathrm{reg}}^{\mathrm{SBL}}(\eta)
-
\overline{\mathcal{L}}_{\mathrm{reg}}^{\mathrm{DFL}}(\eta),
\end{equation}
where a larger positive value indicates that DFL achieves lower ex-post regret under the same imperfect real-time information.

\begin{figure}[!ht]
\centering
\includegraphics[width=0.45\textwidth]{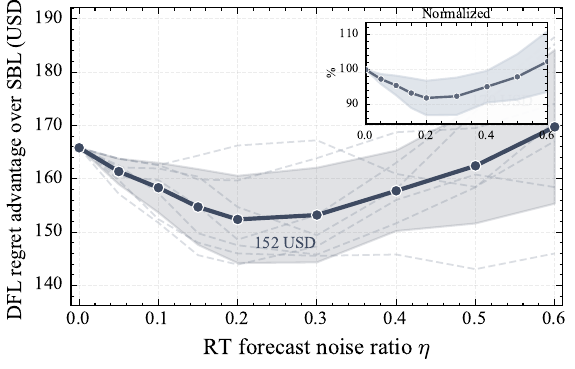}
\caption{Sensitivity of the DFL advantage over SBL to imperfect real-time net-load updates. The solid curve shows the mean advantage $\mathcal{A}(\eta)$ over seven random seeds, the gray dashed curves show seed-wise trajectories, and the shaded band indicates $\pm 1$ standard deviation. The inset reports the normalized advantage $\mathcal{A}(\eta)/\mathcal{A}(0)\times 100\%$.}
\label{fig:rt_sensitivity_advantage}
\end{figure}

Fig.~\ref{fig:rt_sensitivity_advantage} shows that imperfect real-time updates increase the absolute regret of both methods, but do not overturn their relative ranking. At $\eta=0$, DFL reduces the average regret from \$3381.8 to \$3216.0, corresponding to an advantage of \$165.8. As the real-time noise ratio increases to 0.6, the advantage remains positive at \$169.7, and the normalized inset shows that the retained advantage stays within approximately 92\%--102\% of the zero-noise baseline across the entire tested range. In other words, imperfect ultra-short-term updates affect the absolute performance level, but the decision-quality benefit of DFL over SBL remains stable under the same limited-recourse protocol. This result supports the interpretation that the main conclusion of this paper concerns the relative decision effectiveness of the learned day-ahead forecasts, rather than a claim that real-time updates are error-free in deployment.

\subsection{Sensitivity Analysis of the Fisher Regularization}
To further characterize the role of the Fisher-based regularization, a sensitivity analysis is conducted for the coefficient $\lambda$ in \eqref{eq:DFCL_final}. Under the canonical sequence Task~1$\rightarrow$Task~2$\rightarrow\cdots\rightarrow$Task~6, $\lambda$ is varied on a logarithmic grid from $10^{-5}$ to $10^{1}$, while $\lambda=0$ serves as the no-regularization baseline and corresponds to DFL. For each value of $\lambda$, the sequential learning process is repeated over five random seeds under the same network architecture and training configuration. The reported metrics after learning all six tasks include the average regret over the learned tasks.

\begin{figure}[!ht]
\centering
\includegraphics[width=0.48\textwidth]{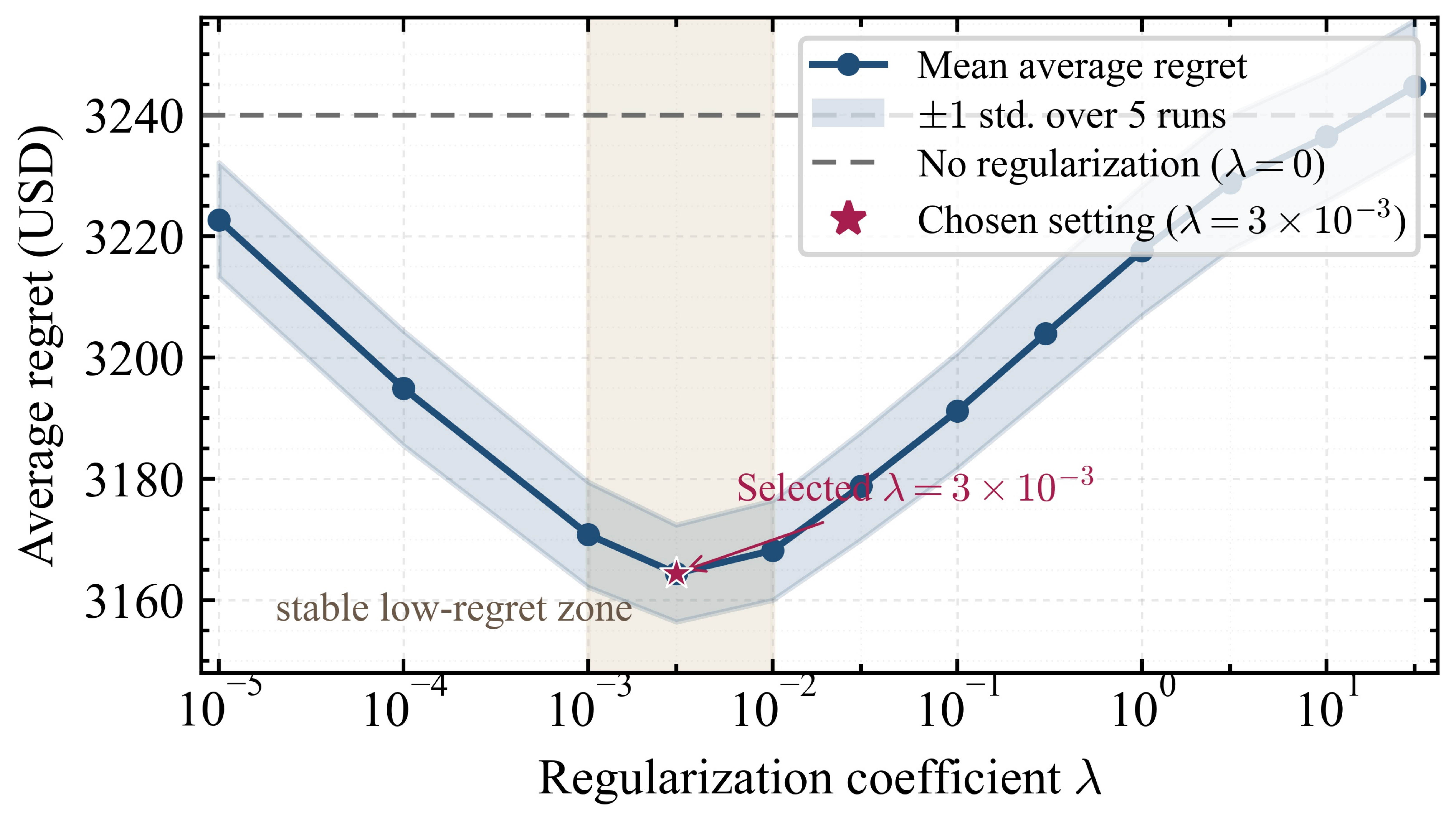}
\caption{Sensitivity of DFCL to the Fisher regularization coefficient $\lambda$. This figure reports the mean average regret after learning all six tasks, with the shaded band indicating $\pm 1$ standard deviation over five runs. }
\label{fig:fim_regularization_sensitivity}
\end{figure}

Fig.~\ref{fig:fim_regularization_sensitivity} exhibits a pronounced U-shaped trend. For small values of $\lambda$, the Fisher penalty provides limited consolidation, and the average regret remains close to the no-regularization baseline. As $\lambda$ increases to approximately $10^{-3}$--$10^{-2}$, all three metrics improve jointly, indicating a more favorable balance between adaptation to incoming tasks and retention of previously acquired decision-relevant information. When $\lambda$ is further increased, the regularization becomes overly restrictive, leading to degraded adaptation on later tasks and a corresponding increase in average regret.

The adopted setting $\lambda=3\times10^{-3}$ lies within this stable low-regret region and yields the lowest mean average regret, approximately \$3164, compared with approximately \$3240 in the no-regularization case. It also reduces RR from approximately 4.85\% to 4.73\% and FM from approximately 311 to 225. Overall, these results indicate that the proposed DFCL remains stable over a moderate range of $\lambda$, while a properly calibrated Fisher regularization materially improves continual decision quality and reduces forgetting.

\begin{table*}[!htbp]
\centering
\small
\caption{Performance comparison of different methods in continual learning}
\label{tab:cl_case_stacked_widths}
\begin{tabular}{>{\raggedright\arraybackslash}p{2.1cm}>{\centering\arraybackslash}p{0.8cm}>{\centering\arraybackslash}p{0.8cm}>{\centering\arraybackslash}p{0.8cm}>{\centering\arraybackslash}p{0.45cm}>{\centering\arraybackslash}p{0.25cm}>{\centering\arraybackslash}p{0.8cm}>{\centering\arraybackslash}p{0.8cm}>{\centering\arraybackslash}p{0.8cm}>{\centering\arraybackslash}p{0.45cm}>{\centering\arraybackslash}p{0.25cm}>{\centering\arraybackslash}p{0.8cm}>{\centering\arraybackslash}p{0.8cm}>{\centering\arraybackslash}p{0.8cm}>{\centering\arraybackslash}p{0.45cm}>{\centering\arraybackslash}p{0.25cm}}
\hline\hline
 & \multicolumn{5}{c}{After Task 1} & \multicolumn{5}{c}{After Task 2} & \multicolumn{5}{c}{After Task 3} \\
\cmidrule(lr){2-6} \cmidrule(lr){7-11} \cmidrule(lr){12-16}
Method & MAE$_{\text{Load}}$ & MAE$_{\text{Price}}$ & Regret & RR & FM & MAE$_{\text{Load}}$ & MAE$_{\text{Price}}$ & Regret & RR & FM & MAE$_{\text{Load}}$ & MAE$_{\text{Price}}$ & Regret & RR & FM \\
\midrule
DFL(Task-wise) & 2.25 & 14.22 & 3106.71 & 4.56\% & 0 & 2.25 & 14.53 & 3035.65 & 4.47\% & 0 & 2.24 & 15.03 & 3036.60 & 4.45\% & 0 \\
SBL & 2.08 & 13.83 & 3305.85 & 4.85\% & 0 & 2.08 & 13.83 & 3264.83 & 4.81\% & 0 & 2.08 & 13.83 & 3260.20 & 4.78\% & 0 \\
DFL(Adam) & 2.25 & 14.22 & 3106.71 & 4.56\% & 0 & 2.27 & 14.14 & 3085.03 & 4.54\% & 91 & 2.21 & 14.59 & 3061.86 & 4.49\% & 110 \\
DFCL(Adam-F) & 2.25 & 14.22 & 3106.71 & 4.56\% & 0 & 2.30 & 13.90 & 3094.49 & 4.56\% & 58 & 2.21 & 14.00 & 3071.34 & 4.50\% & 107 \\
DFCL(EWC) & 2.25 & 14.22 & 3106.71 & 4.56\% & 0 & 2.20 & 13.85 & 3057.84 & 4.50\% & 33 & 2.26 & 14.11 & 3048.45 & 4.47\% & 88 \\
\midrule
 & \multicolumn{5}{c}{After Task 4} & \multicolumn{5}{c}{After Task 5} & \multicolumn{5}{c}{After Task 6} \\
\cmidrule(lr){2-6} \cmidrule(lr){7-11} \cmidrule(lr){12-16}
Method & MAE$_{\text{Load}}$ & MAE$_{\text{Price}}$ & Regret & RR & FM & MAE$_{\text{Load}}$ & MAE$_{\text{Price}}$ & Regret & RR & FM & MAE$_{\text{Load}}$ & MAE$_{\text{Price}}$ & Regret & RR & FM \\
\midrule
DFL(Task-wise) & 2.22 & 15.44 & 3103.36 & 4.59\% & 0 & 2.21 & 15.92 & 3144.23 & 4.68\% & 0 & 2.21 & 16.39 & 3125.79 & 4.68\% & 0 \\
SBL & 2.08 & 13.83 & 3318.80 & 4.91\% & 0 & 2.08 & 13.83 & 3352.31 & 4.99\% & 0 & 2.08 & 13.83 & 3346.18 & 5.01\% & 0 \\
DFL(Adam) & 2.20 & 14.13 & 3173.62 & 4.70\% & 162 & 2.18 & 14.15 & 3208.45 & 4.77\% & 217 & 2.22 & 15.11 & 3240.01 & 4.85\% & 311 \\
DFCL(Adam-F) & 2.15 & 14.19 & 3181.07 & 4.71\% & 174 & 2.17 & 14.10 & 3195.76 & 4.75\% & 203 & 2.14 & 14.15 & 3226.42 & 4.83\% & 278 \\
DFCL(EWC) & 2.26 & 14.21 & 3131.21 & 4.63\% & 101 & 2.19 & 14.15 & 3175.21 & 4.72\% & 171 & 2.20 & 13.98 & 3164.40 & 4.73\% & 225 \\
\hline\hline
\end{tabular}
\end{table*}

\subsection{Computational Cost Analysis}
Additionally, the computational cost of DFCL is compared against decision-focused multi-task learning (DFMTL) \cite{DFMTL} and task-wise DFL under a continual learning setting. Fig.~\ref{fig:training_time} reports the cumulative training time in panel (a) and the memory footprint in panel (b). For DFMTL, the cumulative training time grows non-linearly as tasks accumulate, since new task data are merged with all previous data for joint retraining. Task-wise DFL also exhibits increasing cost because a new model must be trained from scratch for each incoming task. In contrast, DFCL performs regularization-based incremental updates on the current task only, leading to the flattest growth in cumulative training time. By Task~6, the cumulative training time of DFCL is about $4.7\times 10^{3}$~s, compared with about $7.0\times 10^{3}$~s for task-wise DFL and $2.6\times 10^{4}$~s for DFMTL.

Panel (b) addresses the memory scalability of the surrogate buffer. The memory footprint of DFCL stays nearly constant after Task~2 at about 0.12--0.13~GB, whereas task-wise DFL and DFMTL rise to about 0.35~GB and 0.75~GB, respectively, by Task~6. This bounded cost stems from maintaining a single continually updated predictor, diagonal Fisher statistics, and a capped surrogate memory, rather than storing all historical data for joint retraining or keeping one model per task. These results indicate that the proposed online adaptation mechanism remains practical for long task streams in both training time and memory usage.
\begin{figure*}[!ht]
\centering
\includegraphics[width=1\textwidth]{training_time_comparison_bars}
\caption{Comparison of cumulative training time and memory footprint of different methods in the continual learning setting.}
\label{fig:training_time}
\end{figure*}

\section{Conclusion}
\label{sec:6}
This work addresses the task-generalization challenge of decision-focused forecasting in the sequential task stream of PLS scheduling. The proposed DFCL framework avoids retraining a bespoke model for each incoming task and consistently improves decision quality and cross-task generalization over naïve Adam-based continual learning, without requiring enumerating or covering all possible task configurations in advance. These properties make DFCL a practical solution for sustaining high-quality seaport scheduling decisions under evolving operating conditions. The current validation is based on a mixed-source benchmark and therefore mainly shows relative gains under controlled task-structure shifts. Future work will validate DFCL on geographically co-located port datasets, and extend the current limited-recourse formulation to rolling-horizon or multi-stage corrective scheduling with imperfect ultra-short-term forecasts and partial recourse for ESS and logistics decisions. 

{\appendices
\renewcommand{\thefigure}{\Alph{section}.\arabic{figure}}
\renewcommand{\thetable}{\Alph{section}.\arabic{table}}
\renewcommand{\theequation}{\Alph{section}.\arabic{equation}}
\renewcommand{\theHfigure}{appendix.\Alph{section}.\arabic{figure}}
\renewcommand{\theHtable}{appendix.\Alph{section}.\arabic{table}}
\renewcommand{\theHequation}{appendix.\Alph{section}.\arabic{equation}}
\newcommand{\resetappendixcounters}{%
  \setcounter{figure}{0}%
  \setcounter{table}{0}%
  \setcounter{equation}{0}%
}

\section{Details and Operational Constraints of Power-Logistics Scheduling}
\label{sec:appendixA}
\resetappendixcounters
\begin{figure}[!ht]
  \centering
  \includegraphics[width=0.48\textwidth]{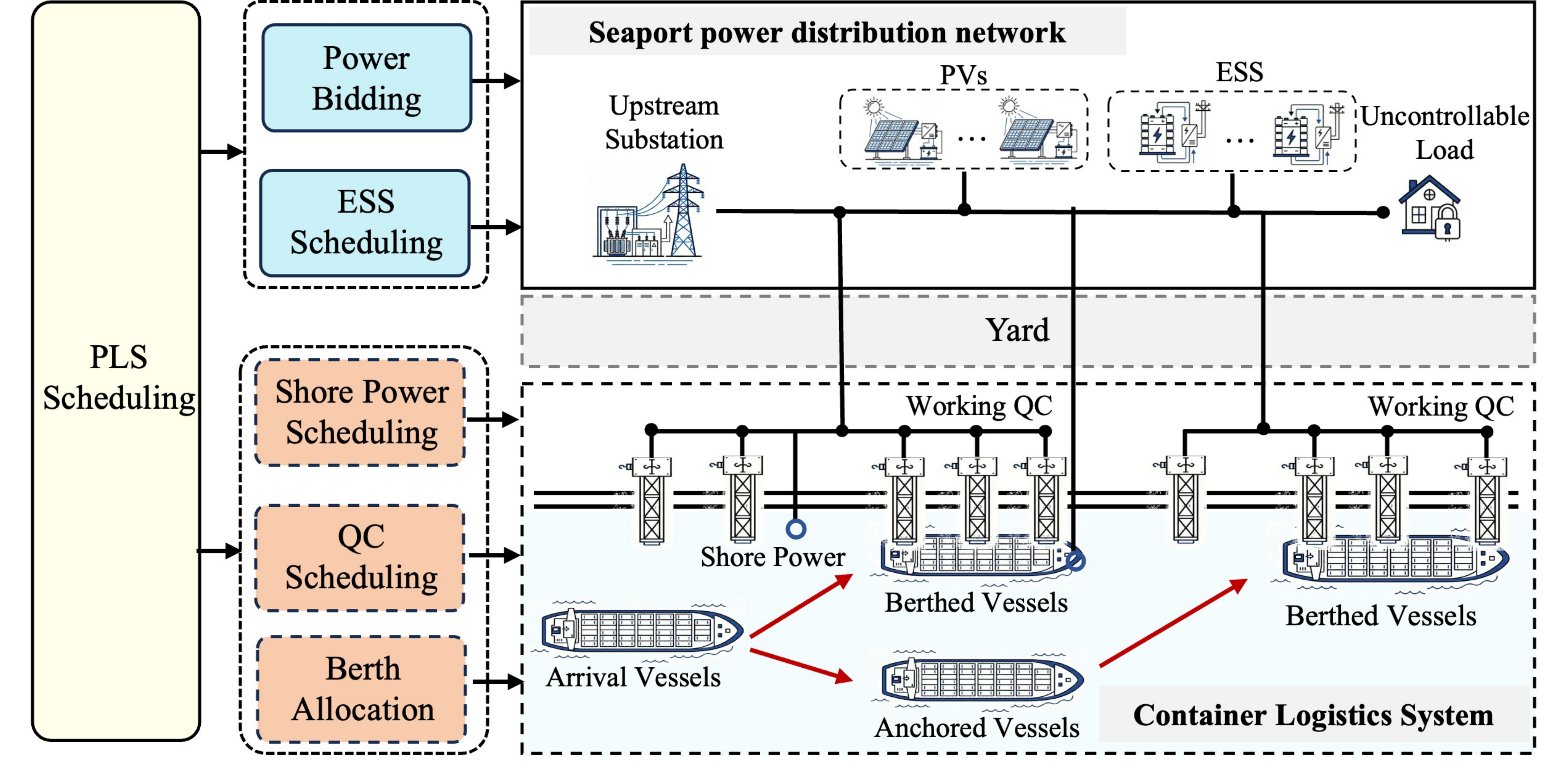}
  \caption{Illustration of the seaport power--logistics system with five coupled scheduling tasks, namely power bidding, ESS scheduling, shore power scheduling, QC scheduling, and berth allocation. The figure shows how the port power network and container logistics operations are jointly coordinated.}
  \label{fig:PLS}
\end{figure}
Fig.~\ref{fig:PLS} illustrates the structure of a typical seaport PLS. The upper layer of Fig.~\ref{fig:PLS} represents the electrical distribution network within the port area. The port is connected to the upstream utility grid through a substation and integrates local energy resources such as PV generation and ESS. These resources supply electricity to various port loads, including uncontrollable facility demand and flexible logistics loads. The lower layer of Fig.~\ref{fig:PLS} depicts the container logistics system. When vessels arrive at the port, they must first be assigned to available berths along the quay. Once a vessel is berthed, quay cranes are allocated to perform container loading and unloading operations. During this process, ships can also receive shore power supply to replace onboard diesel generators. These logistics operations introduce several decision-making tasks that directly influence the electrical load profile and operating cost of the port, including:

\begin{itemize}
\item \textbf{Berth allocation}: determines the spatial position and
time interval during which each vessel occupies the quay.

\item \textbf{Quay crane scheduling}: determines how many cranes are
assigned to each vessel over time, which directly determines crane
electricity consumption.

\item \textbf{Shore power scheduling}: determines the charging power
delivered to berthed vessels during their stay at the port.
\end{itemize}

The constraints in \eqref{eq:bidding2} that characterize the energy storage system (ESS) are expressed as 
\begin{subequations}
\label{eq:ES}
\begin{align}
& E_t - E_{t-1}= \Delta t \left( \eta_{\mathrm{ch}} P_{\mathrm{ES,ch},t}
- \frac{1}{\eta_{\mathrm{dch}}} P_{\mathrm{ES,dch},t} \right), \quad \forall t \geq 1,
\\
& E_{\min} \le\; E_t \;\le E_{\max},\quad \forall  t \geq 0,\\
& 0 \le\; P_{\mathrm{ES,ch},t} \;\le P_{\mathrm{ES}}^{\max},
\quad \forall t \geq 0, \\
& 0 \le\; P_{\mathrm{ES,dch},t} \;\le P_{\mathrm{ES}}^{\max},
\quad \forall t \geq 0 , \\
& E_0 = 50\%E_{\max}.
\end{align}
\end{subequations}

\noindent where \(E_t\) denotes the energy stored in the ESS at time period \(t\); \(P_{\mathrm{ES,ch},t}\) and \(P_{\mathrm{ES,dch},t}\) denote the charging and discharging power at time period \(t\), respectively; \(\Delta t\) is the duration of each scheduling interval; \(\eta_{\mathrm{ch}}\) and \(\eta_{\mathrm{dch}}\) are the charging and discharging efficiencies, respectively; \(E_{\min}\) and \(E_{\max}\) denote the lower and upper energy limits of the ESS; and \(P_{\mathrm{ES}}^{\max}\) is the maximum charging/discharging power.

The constraints in \eqref{eq:bidding3} are detailed as follows. The nonlinear terms in \eqref{eq:bidding3} can be equivalently reformulated into a set of mixed-integer linear constraints using the big-\(M\) method. The detailed reformulation is omitted for brevity, and interested readers are referred to \cite{PLS_sh_4}.

\paragraph*{Ship Scheduling}  
Let $t_j^{\mathrm{in}}$ and $t_j^{\mathrm{out}}$ denote the berthing time and actual departure time of vessel $j$, respectively, and let $v_{jt}$ be an entry of the matrix $\mathbf{V}\in \{0,1\}^{|\mathcal{J}|\times T}$, which indicates whether vessel $j$ is at berth at time $t$. Then
\begin{subequations}
\begin{align}
    & t_j^{\mathrm{arr}} \leq t_j^{\mathrm{in}} \leq T, \quad \forall j, \\
    & t_j^{\mathrm{in}} + \frac{Q_j}{\Delta t \, \eta_{\mathrm{qc}} C_{\mathrm{qc},j}^{\max}} \leq t_j^{\mathrm{out}} \leq t_j^{\mathrm{in}} + \frac{Q_j}{\Delta t \, \eta_{\mathrm{qc}} C_{\mathrm{qc},j}^{\min}} \quad \forall j, \\
    & t_j^{\mathrm{out}} \leq t_j^{\max}, \quad t_j^{\mathrm{in}} - t_j^{\mathrm{arr}} \leq t_j^{\mathrm{wait}} \quad \forall j, \\
    & v_{jt} = \mathbf{1} \left\{\, t_j^{\mathrm{in}} \leq t \leq t_j^{\mathrm{out}} \,\right\}, \quad \forall j,t, \\
    & v_{jt} \geq v_{j,t+1}, \quad \forall j,\, t_j^{\mathrm{in}} \leq t < T,
\end{align}
\end{subequations}
where $t_j^{\mathrm{arr}}$ is the arrival time of vessel $j$, and $Q_j$, $t_j^{\max}$, $t_j^{\mathrm{wait}}$, $\Delta t$, $\eta_{\mathrm{qc}}$, $C_{\mathrm{qc},j}^{\min}$, and $C_{\mathrm{qc},j}^{\max}$ denote the cargo volume, the latest allowable departure time, the maximum waiting time, the scheduling interval, the quay-crane handling efficiency, and the minimum and maximum numbers of quay cranes that can be assigned to vessel $j$, respectively.

\paragraph*{Quay Crane Assignment}  
Quay cranes are the primary cargo-handling equipment in container ports. Their assignment determines both the cargo handling rate and the electricity consumption of crane operations. For each time $t$, let $c_{kjt}$ denote the binary assignment of crane $k \in \mathcal{K}$ to vessel $j$. Then
\begin{subequations}
\begin{align}
    & C_{\mathrm{qc},j}^{\min} v_{jt} \leq \sum_{k \in \mathcal{K}} c_{kjt} \leq C_{\mathrm{qc},j}^{\max} v_{jt}, \\
    & \sum_{j \in \mathcal{J}} c_{kjt} \leq 1, \quad \forall k,t, \\
    & c_{k+1,jt} + c_{k-1,jt} - c_{kjt} \leq 1, \quad \forall j,k,t.
\end{align}
\end{subequations}
\paragraph*{Berth Assignment}  
Let $b_j$ be the berth position of vessel $j$ and $l_j$ its length. Let $\mathcal{S}_i=[b_i, b_i + l_i]$ and $\tau_i=[t_i^{\mathrm{in}}, t_i^{\mathrm{out}}]$ be the spatial and temporal occupation intervals of vessel $i$. Then
\begin{subequations}
\begin{align}
    & b_j \geq 0, \quad b_j + l_j \leq L_{\text{tot}}, \\
    & (\mathcal{S}_i \cap \mathcal{S}_j = \emptyset) \vee (\tau_i \cap \tau_j = \emptyset), \quad \forall i \neq j,
\end{align}
\end{subequations}
where $L_{\text{tot}}$ is the total length of the berth.
\paragraph*{Load and Power Constraints}
Define the total number of cranes in operation at time $t$ as
$\sigma_t := \mathbf{1}_{|\mathcal{K}|}^{\top}\mathbf{C}_t\mathbf{1}_{|\mathcal{J}|}$, where $\mathbf{1}_n$ denotes the $n$-dimensional all-ones vector, $\mathbf{C}_t\in\{0,1\}^{|\mathcal{K}|\times |\mathcal{J}|}$, and $(\mathbf{C}_t)_{kj} = c_{kjt}$. Stacking across all time periods, we obtain
\begin{subequations}
\begin{equation}
\begin{aligned}
    \mathbf{P}_{\mathrm{QC}}
    &= P_{\mathrm{qc}}^{\mathrm{rated}}
    \begin{bmatrix}
        \mathbf{1}_{|\mathcal{K}|}^{\top}\mathbf{C}_1\mathbf{1}_{|\mathcal{J}|} \\
        \vdots \\
        \mathbf{1}_{|\mathcal{K}|}^{\top}\mathbf{C}_T\mathbf{1}_{|\mathcal{J}|}
    \end{bmatrix}
    \in \mathbb{R}^{T},
\end{aligned}
\end{equation}
\begin{equation}
\begin{aligned}
  \label{eq:shore_power}
    \mathbf{P}_{\mathrm{s}} &= (\mathbf{P}^{\mathrm{base}}\mathbf{1}_T^\top+\mathbf{P}^{\mathrm{chg}}) \odot \mathbf{V},
\end{aligned}
\end{equation}
\begin{equation}
\begin{aligned}
    \mathbf{0} &\leq \mathbf{P}^{\mathrm{chg}} \leq (\mathbf{P}^{\mathrm{chg},\max}\mathbf{1}_T^\top)\odot \mathbf{V},
\end{aligned}
\end{equation}
\begin{equation}
\begin{aligned}
    \mathbf{P}^{\mathrm{chg}}\mathbf{1}_T &\geq \mathbf{E}^{\mathrm{chg}},
\end{aligned}
\end{equation}
\end{subequations}
where parameter $P_{\mathrm{qc}}^{\mathrm{rated}}$ is the rated power of a single quay crane, and $\mathbf{P}^{\mathrm{base}}$, $\mathbf{P}^{\mathrm{chg},\max}$, and $\mathbf{E}^{\mathrm{chg}} \in \mathbb{R}^{|\mathcal{J}|}$ are the base load vector, the maximum charging-power vector, and the total charging-demand vector of all vessels, respectively. The variable $\mathbf{P}^{\mathrm{chg}}\in\mathbb{R}_{\ge 0}^{|\mathcal{J}|\times T}$ is the charging-power matrix of all vessels, where each row corresponds to a vessel and each column to a time period. The operator $\odot$ denotes the Hadamard product.

\section{Justification of the Precision Matrix  Approximation}
\label{sec:appendixB}
\resetappendixcounters

The precision matrix $\mathbf{\Lambda}_{k-1}$, defined as the Hessian of the negative log-posterior at $\varTheta_{k-1}^{\ast}$, can be decomposed as
\begin{equation}
\mathbf{\Lambda}_{k-1}
~=~
\Big[\nabla_{\varTheta}^{2}\!\big(-\log \mathbb{P}(\varTheta)-\sum_{t=1}^{k-1}\log \mathbb{P}(\mathcal{D}_t\mid\varTheta)\big)\Big]_{\varTheta_{k-1}^{\ast}}.
\end{equation}
In practice, the prior curvature can be absorbed into a constant quadratic regularizer, leading to the approximation
\begin{equation}
\label{eq:precision_approx}
\mathbf{\Lambda}_{k-1}
~\approx~
\mathbf{H}_{k-1}
~=~
\sum_{t=1}^{k-1}
\Big[\nabla_{\varTheta}^{2}\big(-\log \mathbb{P}(\mathcal{D}_t\mid\varTheta)\big)\Big]_{\varTheta=\varTheta_{k-1}^{\ast}}.
\end{equation}

To relate $\mathbf{H}_{k-1}$ to a computable quantity, we recall the classical Fisher identity, which holds for negative log-likelihood objectives.

\textbf{Lemma 1 (Fisher Identity).}
Let $\mathbb{P}(z\mid\varTheta)$ be a correctly specified statistical model, and assume the usual regularity conditions so that differentiation and integration can be interchanged.
Denote $s(z;\varTheta)\triangleq\nabla_{\varTheta}\log \mathbb{P}(z\mid\varTheta)$. Then
\begin{equation}
\label{eq:fisher_identity}
\mathbb{E}_{z\sim \mathbb{P}(\cdot\mid\varTheta)}\!\big[s(z;\varTheta)\,s(z;\varTheta)^\top\big]
\;=\;
-\;\mathbb{E}_{z\sim \mathbb{P}(\cdot\mid\varTheta)}\!\big[\nabla_{\varTheta}^{2}\log \mathbb{P}(z\mid\varTheta)\big].
\end{equation}

In our DFCL formulation, the per-sample training signal is the regret loss $\ell(z;\varTheta)$ (i.e., decision regret), which is not a log-likelihood in general.
Therefore, the use of a Fisher-type matrix for regret should be understood through Assumption~1 in the main text rather than as a classical likelihood-based FIM. For completeness, we restate its density-level form and derive the resulting positive semidefinite curvature surrogate.

\textbf{Assumption 1 (Regret-induced pseudo-likelihood; repeated).}
For each task $t$, we associate $\ell(z;\varTheta)$ with a Gibbs/energy-based pseudo-likelihood
\begin{equation}
\label{eq:pseudo_likelihood}
\begin{aligned}
\tilde{\mathbb{P}}_t(z\mid \varTheta)
~\triangleq~
\frac{\exp\!\big(-\ell(z;\varTheta)/T\big)}{\tilde{Z}_t(\varTheta)},
\\
\tilde{Z}_t(\varTheta)=\int \exp\!\big(-\ell(\xi;\varTheta)/T\big)\,d\xi,
\end{aligned}
\end{equation}
where $T>0$ is a temperature parameter.
This construction encodes the monotonic relation that parameter regions yielding larger regret correspond to lower likelihood under $\tilde{\mathbb{P}}_t$.

Under \eqref{eq:pseudo_likelihood}, the score function is
\begin{equation}
\label{eq:score_pseudo}
\nabla_{\varTheta}\log \tilde{\mathbb{P}}_t(z\mid \varTheta)
~=~
-\frac{1}{T}\nabla_{\varTheta}\ell(z;\varTheta)
\;-\;
\nabla_{\varTheta}\log \tilde{Z}_t(\varTheta).
\end{equation}
Moreover, by differentiating $\tilde{Z}_t(\varTheta)$ under the integral sign,
\begin{equation}
\label{eq:grad_logZ}
\nabla_{\varTheta}\log \tilde{Z}_t(\varTheta)
~=~
-\frac{1}{T}\,
\mathbb{E}_{z\sim \tilde{\mathbb{P}}_t(\cdot\mid\varTheta)}
\!\big[\nabla_{\varTheta}\ell(z;\varTheta)\big].
\end{equation}
Substituting \eqref{eq:grad_logZ} into \eqref{eq:score_pseudo} yields the centered form
\begin{equation}
\label{eq:score_centered}
\nabla_{\varTheta}\log \tilde{\mathbb{P}}_t(z\mid \varTheta)
~=~
-\frac{1}{T}
\Big(\nabla_{\varTheta}\ell(z;\varTheta)
-\mathbb{E}_{\tilde{\mathbb{P}}_t}\!\big[\nabla_{\varTheta}\ell(z;\varTheta)\big]\Big).
\end{equation}

Therefore, the Fisher information matrix of the pseudo-likelihood satisfies
\begin{equation}
\label{eq:fisher_cov_grad}
\begin{aligned}
  \tilde{\mathbf{F}}_t(\varTheta)
~\triangleq~
\mathbb{E}_{z\sim \tilde{\mathbb{P}}_t(\cdot\mid\varTheta)}
\!\Big[
\nabla_{\varTheta}\log \tilde{\mathbb{P}}_t(z\mid \varTheta)\,
\nabla_{\varTheta}\log \tilde{\mathbb{P}}_t(z\mid \varTheta)^\top
\Big]
\\
~=~
\frac{1}{T^{2}}\,
\mathrm{Cov}_{z\sim \tilde{\mathbb{P}}_t(\cdot\mid\varTheta)}
\!\Big(\nabla_{\varTheta}\ell(z;\varTheta)\Big),
\end{aligned}
\end{equation}
which is positive semidefinite by construction.
The constant factor $1/T^2$ can be absorbed into the regularization weight in the main objective.

In implementation, we estimate curvature from finite samples in $\mathcal{D}_t$.
When $\varTheta$ is close to a stationary point for task $t$ (as is the case for $\varTheta_{k-1}^{\ast}$ after learning tasks $1{:}(k-1)$), the mean gradient is small and the covariance in \eqref{eq:fisher_cov_grad} is well-approximated by the second moment of per-sample gradients.
Accordingly, we use the empirical outer-product estimator
\begin{equation}
\label{eq:ggn_proxy}
\widehat{\mathbf{F}}_t(\varTheta)
~=~
\frac{1}{N_t}\sum_{n=1}^{N_t}
\nabla_{\varTheta}\ell(z^{(t)}_n;\varTheta)\,
\nabla_{\varTheta}\ell(z^{(t)}_n;\varTheta)^\top,
\end{equation}
which corresponds to the empirical Fisher (outer product of scores) under the pseudo-likelihood interpretation and serves as a Fisher-like/generalized Gauss--Newton curvature surrogate for constructing \eqref{eq:precision_approx}.
This yields Eq.~\eqref{eq:emp_fisher} in the main text.

\section{Soft-top-k Selection of Surrogate Buffer Samples}
\label{sec:appendixC}
\resetappendixcounters

Let the surrogate buffer be
\(
\mathcal{M}
=
\{(\mathbf{u}^{(i)},\mathbf{P}_{\mathrm{QC}}^{(i)},\mathbf{V}^{(i)})\}_{i=1}^{M}
\),
and let $\widehat{\mathbf{u}}$ denote the current query. Define the similarity scores
$s_i=\mathrm{sim}(\widehat{\mathbf{u}},\mathbf{u}^{(i)})$ for $i=1,\ldots,M$.
Hard top-$K$ neighbor selection is not suitable for end-to-end training because the active index set changes discontinuously when two similarities exchange their order.
We therefore replace the binary selection rule by the smooth step function
\begin{equation}
\phi(t)
=
\begin{cases}
1-\dfrac{1}{2}e^{-t}, & t\ge 0,\\[4pt]
\dfrac{1}{2}e^{t}, & t<0,
\end{cases}
\label{eq:soft_topk_step}
\end{equation}
and define the soft mask as
\begin{equation}
m_i=\phi(s_i-\lambda^\star),\qquad
\sum_{i=1}^{M} m_i=K.
\label{eq:soft_topk_mask}
\end{equation}
Hence, the hard threshold is replaced by a smooth threshold $\lambda^\star$ whose value is chosen such that the total selection mass equals $K$.

To compute $\lambda^\star$ stably, the similarities are sorted as
$s_{(1)}\le \cdots \le s_{(M)}$.
On each interval $[s_{(r)},s_{(r+1)})$, the equation in \eqref{eq:soft_topk_mask} admits a closed-form solution that depends on cumulative prefix/suffix log-sum-exp terms,
\begin{equation}
\lambda_r^\star
=
\ell_r^{(1)}-\log\!\Big(
\sqrt{\Delta_r^2+\exp(\ell_r^{(1)}+\ell_r^{(2)})}
+\Delta_r
\Big),
\label{eq:soft_topk_lambda}
\end{equation}
where
\(
\ell_r^{(1)}=\log\sum_{j\le r}e^{s_{(j)}}
\),
\(
\ell_r^{(2)}=\log\sum_{j>r}e^{-s_{(j)}}
\),
and
\(
\Delta_r=K-(M-1-r)
\).
The valid threshold is the one whose value lies in its corresponding interval.
The final aggregation weights are then
\begin{equation}
\begin{aligned}
\alpha_i
=
\frac{m_i}{\sum_{j=1}^{M}m_j+\varepsilon},
\\
(\hat{\mathbf{P}}_{\mathrm{QC}},\hat{\mathbf{V}})(\widehat{\mathbf{u}})
=
\sum_{i=1}^{M}\alpha_i(\mathbf{P}_{\mathrm{QC}}^{(i)},\mathbf{V}^{(i)}).
\end{aligned}
\label{eq:soft_topk_appendix_agg}
\end{equation}
For a fixed ordering of $\{s_i\}$, \eqref{eq:soft_topk_step}--\eqref{eq:soft_topk_appendix_agg} are compositions of smooth functions, so gradients can be back-propagated almost everywhere through the surrogate buffer selection step.
When the similarities are well separated, $m_i$ approaches $\{0,1\}$ and the operator recovers the behavior of hard top-$K$ selection.

\section{Additional Details of the Case Study}
\label{sec:appendixD}
\resetappendixcounters

The six tasks in this case study are derived from the vessel arrival and port operation dataset, with each task corresponding to a distinct scheduling scenario characterized by different vessel arrival patterns, cargo demands, and operational constraints. Table~\ref{tab:summary_statistics} summarizes the vessel-specific scheduling parameters for six representative tasks. These tasks are designed to reflect a range of operational conditions, including variations in vessel traffic intensity, cargo volume, and power demand, thereby providing a comprehensive testbed for evaluating the continual learning performance of the proposed DFCL framework in seaport power-logistics scheduling.

\begin{table*}[!ht]
\centering
\caption{Vessel-specific scheduling parameters for six representative tasks.}
\label{tab:summary_statistics}
\scriptsize
\setlength{\tabcolsep}{2.2pt}
\renewcommand{\arraystretch}{1.05}
\resizebox{\textwidth}{!}{%
\begin{tabular}{*{11}{c}!{\hspace{3mm}\vrule width 0.5pt\hspace{3mm}}*{11}{c}}
  
\hline\hline
\multicolumn{11}{c}{\textbf{Task 1}} &
\multicolumn{11}{c}{\textbf{Task 2}} \\
\midrule
$j$ & $t_j^{\mathrm{arr}}$ & $t_j^{\max}$ & $Q_j$ & $C_{\mathrm{qc},j}^{\min}$ & $C_{\mathrm{qc},j}^{\max}$ & $P_j^{\mathrm{base}}$ & $E_j^{\mathrm{chg}}$ & $P_j^{\mathrm{chg},\max}$ & $l_j$ & $t_j^{\mathrm{wait}}$ & $j$ & $t_j^{\mathrm{arr}}$ & $t_j^{\max}$ & $Q_j$ & $C_{\mathrm{qc},j}^{\min}$ & $C_{\mathrm{qc},j}^{\max}$ & $P_j^{\mathrm{base}}$ & $E_j^{\mathrm{chg}}$ & $P_j^{\mathrm{chg},\max}$ & $l_j$ & $t_j^{\mathrm{wait}}$ \\
\midrule
1 & 0.0  & 14.0 & 2866.0 & 1.0 & 3.0 & 4.0 & 7.0  & 1.59  & 148.0 & 5.0 &
1 & 0.0  & 14.0 & 2866.0 & 1.0 & 3.0 & 3.0 & 12.0 & 1.36  & 148.0 & 5.0 \\
2 & 0.0  & 9.0  & 1816.0 & 1.0 & 3.0 & 1.0 & 11.0 & 2.50  & 122.0 & 5.0 &
2 & 3.0  & 12.0 & 606.0  & 1.0 & 1.0 & 3.0 & 14.0 & 2.46  & 143.0 & 5.0 \\
3 & 0.0  & 14.0 & 913.0  & 1.0 & 3.0 & 1.0 & 0.0  & 0.00  & 164.0 & 5.0 &
3 & 4.0  & 10.0 & 869.0  & 2.0 & 3.0 & 3.0 & 0.0  & 0.00  & 151.0 & 5.0 \\
4 & 2.0  & 10.0 & 1606.0 & 1.0 & 3.0 & 4.0 & 11.0 & 1.375 & 199.0 & 5.0 &
4 & 6.0  & 12.0 & 549.0  & 1.0 & 3.0 & 2.0 & 0.0  & 0.00  & 113.0 & 5.0 \\
5 & 3.0  & 9.0  & 493.0  & 1.0 & 2.0 & 2.0 & 8.0  & 1.66  & 95.0  & 5.0 &
5 & 8.0  & 14.0 & 846.0  & 2.0 & 3.0 & 2.0 & 0.0  & 0.00  & 121.0 & 5.0 \\
6 & 5.0  & 15.0 & 1351.0 & 1.0 & 2.0 & 2.0 & 11.0 & 2.16  & 134.0 & 5.0 &
6 & 9.0  & 17.0 & 1071.0 & 2.0 & 2.0 & 3.0 & 13.0 & 3.33  & 111.0 & 5.0 \\
7 & 7.0  & 20.0 & 746.0  & 1.0 & 1.0 & 2.0 & 0.0  & 0.00  & 102.0 & 5.0 &
7 & 10.0 & 17.0 & 931.0  & 1.0 & 2.0 & 3.0 & 6.0  & 1.25  & 145.0 & 5.0 \\
8 & 10.0 & 22.0 & 2446.0 & 1.0 & 3.0 & 3.0 & 0.0  & 0.00  & 83.0  & 5.0 &
8 & 11.0 & 24.0 & 886.0  & 1.0 & 1.0 & 3.0 & 9.0  & 3.75  & 88.0  & 5.0 \\
9 & 14.0 & 26.0 & 668.0  & 1.0 & 3.0 & 1.0 & 0.0  & 0.00  & 107.0 & 5.0 &
9 & 15.0 & 21.0 & 567.0  & 1.0 & 3.0 & 1.0 & 0.0  & 0.00  & 184.0 & 5.0 \\

\hline\hline
\multicolumn{11}{c}{\textbf{Task 3}} &
\multicolumn{11}{c}{\textbf{Task 4}} \\
\midrule
$j$ & $t_j^{\mathrm{arr}}$ & $t_j^{\max}$ & $Q_j$ & $C_{\mathrm{qc},j}^{\min}$ & $C_{\mathrm{qc},j}^{\max}$ & $P_j^{\mathrm{base}}$ & $E_j^{\mathrm{chg}}$ & $P_j^{\mathrm{chg},\max}$ & $l_j$ & $t_j^{\mathrm{wait}}$ & $j$ & $t_j^{\mathrm{arr}}$ & $t_j^{\max}$ & $Q_j$ & $C_{\mathrm{qc},j}^{\min}$ & $C_{\mathrm{qc},j}^{\max}$ & $P_j^{\mathrm{base}}$ & $E_j^{\mathrm{chg}}$ & $P_j^{\mathrm{chg},\max}$ & $l_j$ & $t_j^{\mathrm{wait}}$ \\
\midrule

1  & 0.0  & 11.0 & 2236.0 & 1.0 & 3.0 & 3.0 & 0.0  & 0.00 & 152.0 & 5.0 &
1  & 0.0  & 13.0 & 558.0  & 1.0 & 3.0 & 1.0 & 0.0  & 0.00 & 166.0 & 5.0 \\
2  & 3.0  & 15.0 & 2446.0 & 1.0 & 3.0 & 1.0 & 13.0 & 2.28 & 188.0 & 5.0 &
2  & 0.0  & 8.0  & 1606.0 & 2.0 & 3.0 & 2.0 & 10.0 & 2.27 & 155.0 & 5.0 \\
3  & 7.0  & 19.0 & 832.0  & 1.0 & 2.0 & 4.0 & 0.0  & 0.00 & 153.0 & 5.0 &
3  & 0.0  & 10.0 & 2026.0 & 1.0 & 3.0 & 2.0 & 12.0 & 1.82 & 169.0 & 5.0 \\
4  & 9.0  & 23.0 & 2866.0 & 1.0 & 3.0 & 4.0 & 5.0  & 1.28 & 153.0 & 5.0 &
4  & 3.0  & 9.0  & 932.0  & 2.0 & 4.0 & 4.0 & 0.0  & 0.00 & 82.0  & 5.0 \\
5  & 10.0 & 20.0 & 1351.0 & 1.0 & 2.0 & 4.0 & 0.0  & 0.00 & 144.0 & 5.0 &
5  & 6.0  & 18.0 & 2446.0 & 1.0 & 3.0 & 4.0 & 0.0  & 0.00 & 178.0 & 5.0 \\
6  & 12.0 & 18.0 & 1582.0 & 2.0 & 4.0 & 3.0 & 5.0  & 1.25 & 101.0 & 5.0 &
6  & 9.0  & 19.0 & 466.0  & 1.0 & 3.0 & 4.0 & 0.0  & 0.00 & 99.0  & 5.0 \\
7  & 13.0 & 20.0 & 838.0  & 2.0 & 2.0 & 2.0 & 0.0  & 0.00 & 89.0  & 5.0 &
7  & 10.0 & 23.0 & 582.0  & 1.0 & 3.0 & 2.0 & 0.0  & 0.00 & 147.0 & 5.0 \\
8  & 14.0 & 23.0 & 497.0  & 1.0 & 1.0 & 2.0 & 0.0  & 0.00 & 124.0 & 5.0 &
8  & 13.0 & 26.0 & 2656.0 & 1.0 & 3.0 & 2.0 & 0.0  & 0.00 & 140.0 & 5.0 \\
9  & 15.0 & 25.0 & 673.0  & 1.0 & 3.0 & 2.0 & 0.0  & 0.00 & 110.0 & 5.0 &
9  & 15.0 & 29.0 & 2866.0 & 1.0 & 3.0 & 3.0 & 0.0  & 0.00 & 153.0 & 5.0 \\
  &      &      &        &     &     &     &      &      &     &     &
10 & 16.0 & 28.0 & 681.0  & 1.0 & 2.0 & 4.0 & 0.0  & 0.00 & 124.0 & 5.0 \\
  &      &      &        &     &     &     &      &      &     &     &
11 & 17.0 & 23.0 & 763.0  & 2.0 & 2.0 & 4.0 & 0.0  & 0.00 & 141.0 & 5.0 \\

\hline\hline
\multicolumn{11}{c}{\textbf{Task 5}} &
\multicolumn{11}{c}{\textbf{Task 6}} \\
\midrule
$j$ & $t_j^{\mathrm{arr}}$ & $t_j^{\max}$ & $Q_j$ & $C_{\mathrm{qc},j}^{\min}$ & $C_{\mathrm{qc},j}^{\max}$ & $P_j^{\mathrm{base}}$ & $E_j^{\mathrm{chg}}$ & $P_j^{\mathrm{chg},\max}$ & $l_j$ & $t_j^{\mathrm{wait}}$ & $j$ & $t_j^{\mathrm{arr}}$ & $t_j^{\max}$ & $Q_j$ & $C_{\mathrm{qc},j}^{\min}$ & $C_{\mathrm{qc},j}^{\max}$ & $P_j^{\mathrm{base}}$ & $E_j^{\mathrm{chg}}$ & $P_j^{\mathrm{chg},\max}$ & $l_j$ & $t_j^{\mathrm{wait}}$ \\
\midrule
1 & 0.0  & 7.0  & 931.0  & 1.0 & 2.0 & 1.0 & 0.0  & 0.00 & 83.0  & 5.0 &
1 & 0.0  & 8.0  & 555.0  & 1.0 & 2.0 & 3.0 & 0.0  & 0.00 & 87.0  & 5.0 \\
2 & 0.0  & 7.0  & 450.0  & 1.0 & 1.0 & 1.0 & 0.0  & 0.00 & 173.0 & 5.0 &
2 & 0.0  & 11.0 & 921.0  & 1.0 & 3.0 & 2.0 & 0.0  & 0.00 & 172.0 & 5.0 \\
3 & 1.0  & 9.0  & 905.0  & 2.0 & 4.0 & 2.0 & 0.0  & 0.00 & 198.0 & 5.0 &
3 & 2.0  & 8.0  & 830.0  & 2.0 & 3.0 & 3.0 & 0.0  & 0.00 & 180.0 & 5.0 \\
4 & 4.0  & 17.0 & 1771.0 & 1.0 & 2.0 & 4.0 & 0.0  & 0.00 & 151.0 & 5.0 &
4 & 6.0  & 16.0 & 484.0  & 1.0 & 2.0 & 3.0 & 0.0  & 0.00 & 159.0 & 5.0 \\
5 & 7.0  & 20.0 & 2656.0 & 1.0 & 3.0 & 2.0 & 0.0  & 0.00 & 153.0 & 5.0 &
5 & 8.0  & 21.0 & 886.0  & 1.0 & 1.0 & 1.0 & 9.0  & 1.07 & 160.0 & 5.0 \\
6 & 11.0 & 23.0 & 2446.0 & 1.0 & 3.0 & 4.0 & 0.0  & 0.00 & 156.0 & 5.0 &
6 & 11.0 & 21.0 & 1351.0 & 1.0 & 2.0 & 4.0 & 13.0 & 2.36 & 195.0 & 5.0 \\
7 & 14.0 & 25.0 & 640.0  & 1.0 & 1.0 & 4.0 & 0.0  & 0.00 & 189.0 & 5.0 &
7 & 13.0 & 27.0 & 1911.0 & 1.0 & 2.0 & 3.0 & 12.0 & 2.31 & 78.0  & 5.0 \\
8 & 15.0 & 29.0 & 516.0  & 1.0 & 3.0 & 1.0 & 0.0  & 0.00 & 171.0 & 5.0 &
8 & 17.0 & 28.0 & 746.0  & 1.0 & 1.0 & 4.0 & 0.0  & 0.00 & 156.0 & 5.0 \\
9 & 16.0 & 25.0 & 606.0  & 1.0 & 1.0 & 3.0 & 0.0  & 0.00 & 80.0  & 5.0 &
9 & 18.0 & 31.0 & 886.0  & 1.0 & 1.0 & 2.0 & 13.0 & 2.71 & 143.0 & 5.0 \\
\hline\hline
\end{tabular}%
}

\vspace{1mm}
\parbox{\textwidth}{\footnotesize
\textit{Note:}
ship\_ID denotes the vessel index in each task;
$t^{\mathrm{arr}}$ denotes the vessel arrival time, and $t^{\max}$ denotes the latest allowable departure time;
$Q$ denotes the cargo volume;
$C_{\mathrm{qc}}^{\min}$ and $C_{\mathrm{qc}}^{\max}$ denote the minimum and maximum numbers of quay cranes assigned to the vessel;
$P^{\mathrm{base}}$ denotes the base shore-power demand;
$E^{\mathrm{chg}}$ denotes the charging energy demand;
$P^{\mathrm{chg},\max}$ denotes the maximum charging power;
$l$ denotes the vessel length;
$t^{\mathrm{wait}}$ denotes the maximum allowable waiting time.
}
\end{table*}

}

\end{document}